\documentclass[lettersize,journal]{IEEEtran}
\pdfoutput=1
\usepackage{amsmath,amsfonts}
\usepackage{array}
\usepackage[caption=false,font=footnotesize,labelfont=rm,textfont=rm]{subfig}
\usepackage{textcomp}
\usepackage{stfloats}
\usepackage{url}
\usepackage{verbatim}
\usepackage{graphicx}
\usepackage{cite}
\usepackage{times}
\usepackage{epsfig}
\usepackage{graphicx}
\usepackage{amsmath}
\usepackage{amssymb}
\usepackage{multirow}
\usepackage{booktabs}
\usepackage{pifont}
\usepackage{authblk}
\usepackage{balance}
\usepackage{bm}
\usepackage{lipsum}
\usepackage[inkscapelatex=false]{svg}
\usepackage{subfig}
\usepackage{xcolor}
\usepackage{colortbl} 
\usepackage[ruled]{algorithm2e}
\usepackage[colorlinks,
            linkcolor=blue,
            anchorcolor=blue,
            citecolor=blue]{hyperref}
\newcommand{\ie}{\textit{i.e.}}
\newcommand{\eg}{\textit{e.g.}}
\newcommand{\reffig}[1]{Fig. \ref{#1}}
\newcommand{\refsec}[1]{Section \ref{#1}}

\makeatletter
\newcommand{\ssymbol}[1]{^{\@fnsymbol{#1}}}
\makeatother

\begin{document}

\title{Learning Generalizable Models via Disentangling Spurious and Enhancing Potential Correlations

\thanks{Na Wang, Jintao Guo, Yinghuan Shi, and Yang Gao are with the National Key Laboratory for Novel Software Technology and the National Institute of Healthcare Data Science, Nanjing University, Nanjing, China, 210023 (e-mail: wangna@smail.nju.edu.cn; guojintao@smail.nju.edu.cn; syh@nju.edu.cn; gaoy@nju.edu.cn).

Lei Qi is with the School of Computer Science and Engineering, Southeast University, and Key Laboratory of New Generation Artificial Intelligence Technology and Its Interdisciplinary Applications (Southeast University), Ministry of Education, Nanjing, China, 211189 (e-mail: qilei@seu.edu.cn).

Na Wang and Lei Qi are the co-first author.

{*}The corresponding author: Yinghuan Shi.

This work was supported by the NSFC Program (62222604, 62206052, 62192783), Jiangsu Natural Science Foundation (BK20210224).}}

\author{Na Wang$\ssymbol{2}$, Lei Qi$\ssymbol{2}$, Jintao Guo, Yinghuan Shi{*}, Yang Gao}
% <-this % stops a space



\maketitle

\begin{abstract}
% 150 - 250
Domain generalization (DG) intends to train a model on multiple source domains to ensure that it can generalize well to an arbitrary unseen target domain. 
% By introducing irrelevant features, spurious correlations cause the model to overly focus on domain-specific features while neglecting those pertinent to the target task, ultimately impairing the ability of the model to generalize across diverse domains.
The acquisition of domain-invariant representations is pivotal for DG as they possess the ability to capture the inherent semantic information of the data, mitigate the influence of domain shift, and enhance the generalization capability of the model.
Adopting multiple perspectives, such as the sample and the feature, proves to be effective. 
The sample perspective facilitates data augmentation through data manipulation techniques, whereas the feature perspective enables the extraction of meaningful generalization features. 
% through representation learning or feature decoupling.
In this paper, we focus on improving the generalization ability of the model by compelling it to acquire domain-invariant representations from both the sample and feature perspectives by disentangling spurious correlations and enhancing potential correlations.
1) From the sample perspective, we develop a frequency restriction module, guiding the model to focus on the relevant correlations between object features and labels, thereby disentangling spurious correlations.  
% We introduce two methods to preserve frequency components, which suppress the backgrounds of images while retaining semantic information as domain-invariant representations, thereby disentangling spurious correlations.
2) From the feature perspective, the simple Tail Interaction module implicitly enhances potential correlations among all samples from all source domains, facilitating the acquisition of domain-invariant representations across multiple domains for the model.
The experimental results show that Convolutional Neural Networks (CNNs) or Multi-Layer Perceptrons (MLPs) with a strong baseline embedded with these two modules can achieve superior results, \eg, an average accuracy of 92.30$\%$ on Digits-DG. 
Source code is available at \href{https://github.com/RubyHoho/DGeneralization}{\textcolor{blue}{https://github.com/RubyHoho/DGeneralization}}.

\end{abstract}


\section{Introduction}
\label{sec:intro}

\begin{figure}[t]
%%\vspace{-0.8cm}
   \setlength{\belowcaptionskip}{-1cm}
   \setlength{\abovecaptionskip}{-0.01cm}
  \centering
\includegraphics[width=1\linewidth]{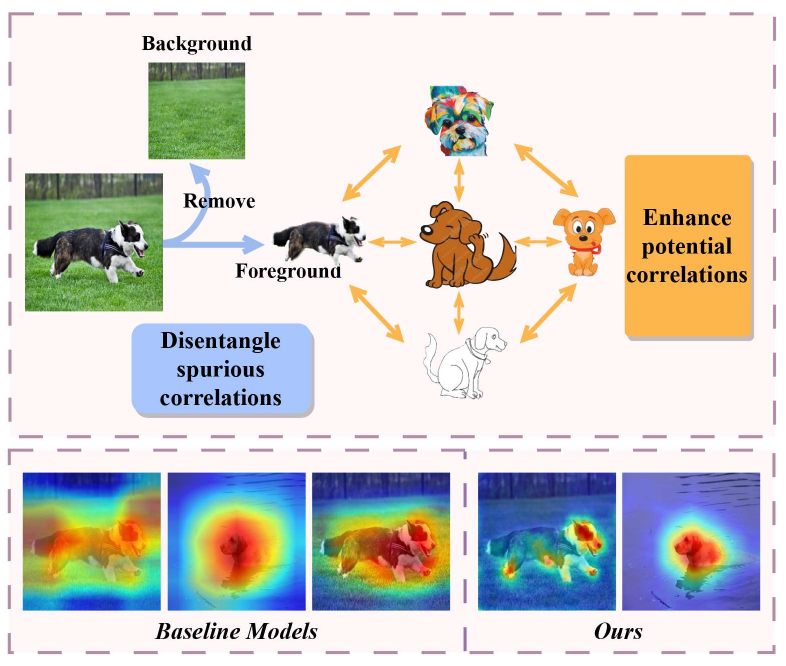}
   \caption{Diagram of disentangling spurious correlations and enhancing potential correlations, aiming to learn domain-invariant representations. 
    The baseline models focus both on objects and backgrounds, while ours focus mainly on the generalized features of the objects.}
   \label{fig:fig1}
\vspace{-10pt}
\end{figure}

\IEEEPARstart{W}{e} have witnessed the remarkable performance improvement achieved by deep models in a wide range of learning tasks under the conventional independent and identically distributed (\textit{i.i.d}) setting. 
However, in real-world applications, the distribution shift during the data acquisition procedure frequently occurs, which significantly violates the \textit{i.i.d} assumption, leading to inferior generalization results.
It is heartening to see increasing research endeavors to resolve this issue, with domain generalization (DG) \cite{40,41} emerging as a popular and promising method.

%meta-learning-based \cite{1,2,3}
Researchers have developed multiple DG methods that roughly fall into data augmentation-based \cite{4,12}, meta-learning-based \cite{2}, contrastive learning-based \cite{12}, and domain-invariant representations learning-based methods \cite{6,8,15,40,83}. 
The goal of DG methods based on domain-invariant representations learning is to extract features that are invariant across domains and can be applied to any other domains with different joint distributions, thus reducing the influence of domain-specific features from the source domains.
One widely used method for achieving this is adversarial training \cite{4,6,8,83}, which seeks to reduce the domain gap between multiple source domains and learn domain-invariant features.
Instead of directly learning domain-invariant features, several methods \cite{85,86} focus on separating domain-specific and domain-invariant features.

Learning domain-invariant features is a challenging task for DG, as different domains exhibit diverse distributions, noises, and biases that affect the feature extraction process.  
Traditional methods may encounter challenges in effectively capturing and utilizing domain-invariant features. 
The difficulty lies in disentangling spurious correlations introduced by domain-specific features and enhancing potential correlations that are truly indicative of the underlying task-relevant information.
Therefore, for encouraging the acquisition of domain-invariant representations, our method aims to tackle these challenges from the creative perspective of correlations, leveraging two complementary strategies achieved via disentangling spurious correlations and enhancing potential correlations.

Spurious correlations typically arise from features irrelevant to the label, such as backgrounds.
The visualized activation regions of vanilla ResNet-18 and MLP models without any DG modules reveal such highly easier-to-fit correlations (\ie, grass as the background is also activated) as shown in \reffig{fig:fig1}. 
Consequently, (1) when encountering images of other objects appearing on the grass, or (2) when encountering changes in the background of the dog, the model is prone to make incorrect predictions.
By disentangling spurious correlations, we eliminate the negative impact of irrelevant features, allowing the model to effectively focus on task-relevant information, \ie, domain-invariant representations. 
This process enhances the discriminative ability of the model, enables it to identify and exclude features unrelated to the target task, and improves its learning ability of the relevant features of the target task.

\begin{figure*}[t]
\vspace{-0.3cm}
   \setlength{\belowcaptionskip}{-1cm}
  \setlength{\abovecaptionskip}{-0.03cm}
  \centering
\includegraphics[width=1\linewidth]{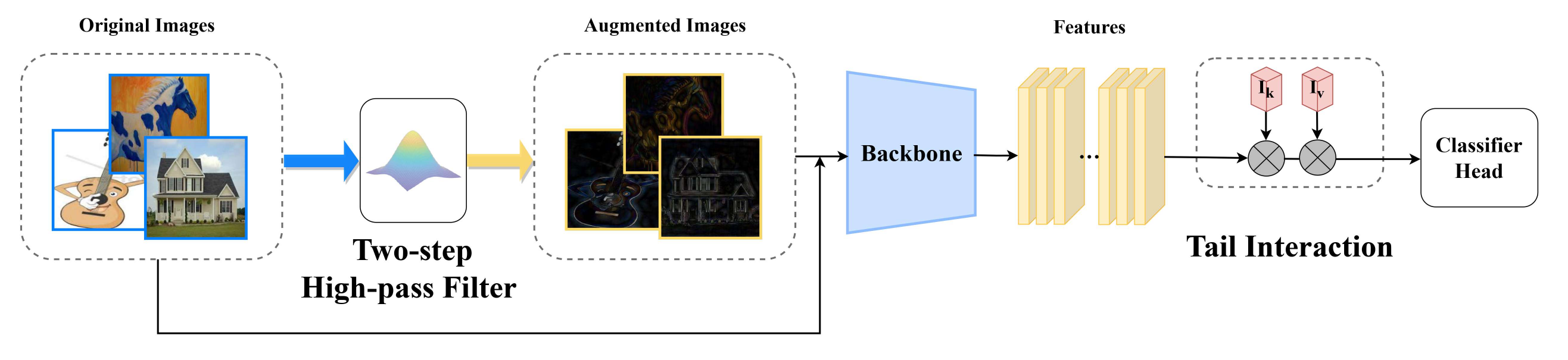}
   \caption{The architecture using our proposed two modules.
   The whole network includes two portable modules: frequency restriction (Gaussian Kernel/Two-step High-pass Filter) and Tail Interaction.
   The network receives both original and augmented images as input.
   At the end of the network, the Tail Interaction is based on two units, $I_k$ and $I_v$, to establish the potential correlations among different samples.}
\label{fig:framework}
\end{figure*}

Potential correlations indicate semantic consistency among different samples, in other words, semantic correlations. 
For instance, there exist potential correlations between dogs from the Cartoon domain and dogs from the Photo domain due to their semantic consistency.
In this paper, by enhancing potential correlations among all samples from all source domains, we aim to capture and amplify the underlying patterns that are informative and indicative of the target task.
These patterns represent hidden interdependencies and correlations that exist across different domains. 
Our method can help the model effectively explore and enhance the utilization of the common features among samples from diverse domains, further boosting the generalization capability \cite{63} of the model in different domains. 
Specifically, from the sample perspective, we introduce a frequency restriction module, and the Tail Interaction module from the feature perspective, as shown in \reffig{fig:framework}. 
This comprehensive consideration enables a more complete feature representation for the model, allowing it to capture a more varied and diverse set of features.
% based on preserving the high frequency components of images

(1) \textit{Frequency restriction}.
Prior studies \cite{66,67} have revealed that the edges and lines of objects are more related to the high frequency components, while the frequency of background and object surface is relatively low \cite{64}.
Consequently, assuming that by learning more from the high frequency components, the model can better extract the semantic concepts of objects while avoiding learning spurious correlations between backgrounds and labels.
% Based on our analysis that the high frequency components can effectively preserve the semantics and disentangle spurious correlations, 
Formally, we impose restrictions in the pixel space or the Fourier spectral space. 
In the pixel space, we transform each image into its high frequency and low frequency components by employing the Gaussian Kernel \cite{78}. 
Besides, we directly filter low frequency in the frequency domain by applying Fast Fourier Transform (FFT) \cite{34} to each image. 
During high-pass filtering, the high frequency component inevitably retains the noise.
Inspired by the fact \cite{69} that an amplitude scaling operation can model the noisy channel, we scale the amplitude and phase components of the Fourier spectrum of the image.
We name the high-pass filter followed by the scaling process ``\textit{Two-step High-pass Filter}".
Moreover, the experiments conducted in \cite{76} have revealed that the model is vulnerable to spurious correlations when there are limited samples available in which such correlations do not exist.
Therefore, we involve the filtered images as augmented images to improve the ability of the model to learn robust features.

(2) \textit{Tail Interaction}.
We further introduce the Tail Interaction module to force the model to enhance potential correlations among all samples from all source domains, acquiring domain-invariant features.
This module computes attention between input features and the interaction unit, implicitly considering the correlations among all samples across multiple domains.
Moreover, the Tail Interaction module offers the flexibility of manual adjustment to achieve much smaller computational complexity.
The Tail Interaction module, with $O(n)$ linear complexity, enables its utilization in high-resolution scenarios.
As shown in \reffig{fig:fig1}, our models mainly focus on features close to the foreground.
% The improved generalization capability of our model based on CNNs or MLPs stems from additional robustness against spurious correlations, which we achieve by incorporating augmented samples that exclude such correlations, and the potential correlations with all samples from various domains. 
% This is accomplished by extracting and aggregating domain-invariant representations, such as the locations and shapes of the target objects.
In summary, our work makes the following contributions.

\begin{itemize}
%, aiming to compel the model to acquire domain-invariant representations.
% \item We address the limitations of traditional methods by explicitly targeting the disentanglement of spurious correlations and the enhancement of potential correlations. 
\item 
% We extensively study the generalization performance of the model from the creative perspective of correlations.
By considering both the sample and feature perspectives, we provide a comprehensive framework for learning domain-invariant representations that can be applied across different domains and achieving improved generalization performance in diverse domains.

\item From the sample perspective, we impose restrictions in the pixel space or the Fourier spectral space to prevent damage from spurious correlations, resulting in the mostly suppressed background of the obtained image while retaining semantics information.
From the feature perspective, we introduce the Tail Interaction module, which enhances the consideration of potential correlations among similar semantic features of objects in different samples from all source domains.

\item Extensive experiments demonstrate that these two modules, whether embedded in the CNN or the MLP model with a strong baseline, can build the new state-of-the-art, \eg, an average accuracy of 92.30$\%$ on Digits-DG based on MLPs.
Additionally, ablation studies validate the efficacy of our proposed modules.
\end{itemize}

The following parts are organized as below. 
\refsec{sec:ralatedWork} briefly reviews the related works. 
The technical details of the proposed method are described in \refsec{sec:method}. 
The experimental comparisons and ablation studies are detailed in \refsec{sec:experiments}. 
Finally, \refsec{sec:conclusion} concludes this paper.
%-------------------------------------------------------------------------
%-------------------------------------------------------------------------

\vspace{-5pt}
\section{Related Work}
\label{sec:ralatedWork}
In this section, we thoroughly review and discuss related works that are directly relevant to our research in the areas of domain generalization, Fourier transform, and dataset bias and spurious correlation mitigation-based methods.

\subsection{Domain Generalization} 
Domain generalization aims to generalize to the unseen domain by training with multiple source domains. 
%\cite{8,13,15,16}
Motivated by domain adaptation methods, initial studies for DG carry out domain alignment \cite{8,15,16} to learn domain-invariant features by reducing the distribution distance among multiple domains. 
Some methods are based on learning domain-invariant representations via kernel methods \cite{40}, domain-adversarial learning \cite{6,8,15}, meta-learning \cite{55,60}, and data augmentation \cite{5,51,52}. 
Besides, person re-identification is reformulated as a multi-dataset domain generalization problem and the designed network is based on an adversarial auto-encoder to learn a generalized domain-invariant latent feature representation with the Maximum Mean Discrepancy (MMD) measure to align the distributions across multiple domains. 
%omains \cite{114}
Learning autonomously allows to discover invariances and regularities that help to generalize.
The model learns the semantic labels in a supervised fashion, and broadens its understanding of the data by learning from self-supervised signals how to solve a jigsaw puzzle on the same images \cite{17}.
A method that iteratively divides samples into latent domains via clustering is proposed in \cite{61}, and which trains the domain invariant feature extractor shared among the divided latent domains via adversarial learning.
Moreover, by building a strong baseline on a lightweight MLP with a filter, researchers propose to perturb local texture features and meanwhile preserve global structure features, thus enabling the filter to remove structure-irrelevant information sufficiently and capture the global representations (\eg, structure) \cite{119}.
Researchers also conduct studies on ensemble learning \cite{19} by training a set of exemplar classifiers with each classifier learnt by using one positive training sample and all negative training samples.
Some studies address practical and challenging scenarios in DG. 
For instance, imbalance domain generalization (IDG) \cite{106} tackles the issue by augmenting reliable samples for minority domains/categories, thereby enhancing the discriminative ability of the learned model. 
Another examples are domain generalization few-shot classification (DG-FSC).
%fsc[110]
%in [111]
The causal invariance of the average causal effect of the features to the labels is considered. This invariance regularizes the training in which interventions are performed on features to enforce stability of the causal prediction by the classifier across domains.
% meta-learning \cite{2,57}
% Some studies focus on data augmentation \cite{5,12,14} using a generator network or Fourier-based augmentation strategy.
Others concentrate on feature adaptation \cite{53}, gradient/neuron diversity \cite{56}, and sample weights \cite{62}.
On the contrary, our method seeks to improve the generalization ability of the model by promoting the extraction of domain-invariant representations from the creative perspective of correlations. 
We achieve this by disentangling spurious correlations and enhancing potential correlations.

% The abovementioned DG methods are CNN-based architectures, and recent works have started investigating the DG performance of the new vision architectures like MLPs \cite{20,21,23,24,58}.
% Nevertheless, the above methods based on ViTs or MLPs ignore the correlations among features, which affects or even impairs the predictions of these new architectures.
% The features learned by the model are not domain-independent and even cannot represent the semantics information correctly.
% Thus they cannot be well generalized to the unseen target domains.
% Different from them, our method overcomes the problem and ensures that both CNNs and MLPs learn domain-invariant representations and avoid suffering from spurious correlations. 

% {\bf MLP-like Vision Models.} 
% % Recently, many researchers have paid much attention to MLP-like 
% Recently, researchers have given significant attention to MLP-like architectures \cite{25,26,27,28} consisting of fully connected layers and non-linear activation functions.
% Although the MLP-like architectures possess simpler architectures and introduce less inductive bias than SOTA models, their performances are still comparable with them.
% Some techniques \cite{29,30} demonstrate that the simple Fourier Transform can be a competitive alternative to either self-attention or MLPs.
% These layers are more efficient and less sensitive to the token length than self-attention and MLPs.

\subsection{Fourier Transform}
Fourier transform is an essential tool in digital image processing for decades. 
With the breakthroughs of CNNs in computer vision \cite{32,33}, various works start incorporating Fourier transform in some deep learning methods \cite{34,36} for vision tasks. %34 35 36,3536
Some of these works \cite{36} employ discrete Fourier transform to convert the images to the frequency domain and leverage the frequency information to improve the performance in specific tasks.
In contrast, others utilize the convolution theorem to accelerate the CNNs via FFT. 
Recent works \cite{5,38} leverage Fourier transform to develop the performance of the deep learning model by generating various augmented images or exchanging information globally among the tokens \cite{30}. 
Some works use regularized Graph Fourier Transform (GFT) coding to accommodate various visible watermarking schemes \cite{112}, and design the Phase Mixing (PM) strategy based on Fourier Transform \cite{113}.
Based on Quasi Fourier-Mellin Transform (QFMT), the Quasi Fourier-Mellin Descriptor (QFMD) is constructed for the extraction of affine invariant features \cite{117}.
Another effort presents a new Light Field representation called Fourier Disparity Layers for efficient Light Field processing and rendering.
%rendering \cite{118}
Our method differs from previous works in that we utilize the information from the Fourier transform of images, including high and low frequencies, amplitude, and phase, to extract complete semantic information with less noise interference as domain-invariant representations.

\subsection{Dataset Bias and Spurious Correlation Mitigation}
Several previous studies \cite{95,96} show that deep networks trained using Empirical Risk Minimization (ERM) tend to exploit biases and spurious correlations present in the training data, leading to poor generalization on test data. 
To address this issue, existing methods focus on various techniques, including re-sampling rare data patterns \cite{97}, adversarial debiasing \cite{101}, model ensembling \cite{102}, minority/counterfactual sample generation \cite{97}, and invariant/robust risk minimization \cite{105}. 
%minimization \cite{77,105}
Many of these methods require the annotation of bias variables or sub-groups within categories \cite{95}. 
Some recent methods have also attempted to detect and mitigate biases without these variables by training separate bias-amplified models for de-biasing the main model \cite{115}.
% model \cite{115,116}
In contrast, we combat dataset biases and spurious correlations from the perspective of generating new unbiased samples based on training data, compensating for limited samples in datasets where spurious correlations are absent.
Additionally, these previous methods did not address this problem from the perspective of correlations between different samples. Conversely, we introduce the Tail Interaction module aimed at learning semantic consistency correlations between samples to mitigate spurious correlations. Being placed at the end of the network, the Tail Interaction module further processes and updates the output features of the network. The module allows the model to focus more on domain-invariant features, improving the performance of domain generalization tasks.
%-------------------------------------------------------------------------
\section{Method}
\label{sec:method}
\subsection{Problem Definition and Overview}
Given $K$ source domains $\mathcal{D}_s=\{\textit{D}^1_s,\textit{D}^2_s,\cdots,\textit{D}^K_s\}$, where $\textit{D}^k_s=\{({x}^k_i, y^k_i)\}^{N_k}_{i=1}$ denotes the \textit{$k$}-th domain consisting of $N_k$ samples ${x}^k_i$ with their corresponding labels ${y^k_i}$, the goal of the DG task is to learn a generalized model from the source domains $\mathcal{D}_s$.
The model can perform well on arbitrary unseen target domains $\mathcal{D}_t$ without requiring additional model updating using samples in $\mathcal{D}_t$.
Put differently, during the testing phase, the model is presented with images from a previously unseen domain $\mathcal{D}_t$ with a distinct joint distribution.

We illustrate an overall framework in \reffig{fig:framework}.
The network contains two vital modules, \ie, (1) frequency restriction (\refsec{sec:gaus}, \refsec{sec:hpf}) and (2) Tail Interaction (\refsec{sec:tail}).
The network processes both original and Two-step High-pass Filtered images, compensating for limited samples in datasets where spurious correlations are absent.
The Tail Interaction module at the end of the network comprises two units, namely $I_k$ and $I_v$, which establish potential correlations among samples from all source domains.

%%\vspace{-20pt}
\subsection{Frequency Extraction via Convolution}
\label{sec:gaus}
Spurious correlations often stem from features that lack relevance to the target label, such as background elements. By disentangling these spurious correlations, we mitigate the adverse effects of irrelevant features, empowering the model to concentrate more effectively on task-relevant information—specifically, domain-invariant representations. This procedure elevates the model's discriminative prowess, enabling it to discern and discard features unrelated to the target task. To disentangle spurious correlations from the sample perspective, we impose restrictions in pixel space to preserve the semantic information of objects.
Some previous DG works \cite{5,68} are motivated by the property of the Fourier transform: the phase components of the Fourier spectrum preserve the high-level semantics of the original signal, while the amplitude components contain the low-level statistics.
In contrast with them, we use another critical property of the Fourier transform: the frequency domain information of the image represents the intensity of gray level change in the image, that is, the gradient of the gray level in the plane space.
The low frequency component of the signal represents the part of the image where the brightness or gray value changes slowly.
In contrast, the high frequency component represents some lines, edges (\eg, areas with sharp pixel changes), and even noise in most cases.

We can transform each image $x$ into the low frequency image $x_L$ and the high frequency image $x_H$.
Specifically, we employ the Gaussian Kernel, which filters the high frequency feature and keeps the low frequency information:
\begin{equation}
\setlength{\abovedisplayskip}{3pt} 
\setlength{\belowdisplayskip}{3pt}
  g_\sigma[p, q]=\frac{1}{2 \pi \sigma^2} e^{-\frac{1}{2}\left(\frac{p^2+q^2}{\sigma^2}\right)},
  \label{eq:g1}
\end{equation}
where $[p,q]$ denotes the spatial location within the image, and $\sigma^2$ denotes the variance of the Gaussian function, respectively. 
The variance is increased proportionally with the Gaussian Kernel size.
We first obtain the low frequency image $x_L$ using the convolution of the Gaussian Kernel:
\begin{equation}
\setlength{\abovedisplayskip}{3pt} 
\setlength{\belowdisplayskip}{3pt}
  x_L[p, q]=\sum_m \sum_n g[m, n] \cdot x[p+m, q+n],
  \label{eq:g2}
\end{equation}
where $m$, $n$ denote the index of a 2D Gaussian Kernel, \ie, $m, n \in\left[-\frac{g-1}{2}, \frac{g-1}{2}\right]$.

To obtain $x_H$, we first convert the color images into grayscale and then subtract the low frequency information:
\begin{equation}
\setlength{\abovedisplayskip}{3pt} 
\setlength{\belowdisplayskip}{3pt}
  x_H=\operatorname{rgb2gray}(x)-(\operatorname{rgb2gray}(x))_L,
  \label{eq:g3}
\end{equation}
where the $\operatorname{rgb2gray()}$ function converts the color image to grayscale, removing the color and illumination information unrelated to the identity and structure.
The resulting high frequency image $x_H$ contains sharp edges.

\subsection{Two-step High-pass Filter}
\label{sec:hpf}
Previous studies \cite{66,67} have indicated that the edges and lines of objects exhibit a stronger association with high frequency components, while the frequency of the background and object surface tends to be relatively low \cite{64}. With this understanding, we hypothesize that by emphasizing learning from high frequency components, the model can more effectively capture the semantic features of objects while mitigating the risk of learning spurious correlations.
In a formal sense, we introduce constraints in the Fourier spectral space to aid the model in eliminating spurious correlations. This involves a direct filtering of low frequencies in the frequency domain using FFT for each image. It is important to note that during high-pass filtering, the high frequency component inevitably retains some level of noise. To mitigate the impact of noise in the image, we implement a scaling operation on both the amplitude and phase components of the Fourier spectrum. This operation is designed to effectively diminish high frequency elements of noise and enhance the overall image quality. This combined process, consisting of high-pass filtering followed by scaling operations, is referred to as the ``\textit{Two-step High-pass Filter}".

\textbf{The First Step: High-pass Filter.}
A high-pass filter is a typical filtering method. 
The basic rule is that high frequency signals can pass through, while low frequency signals below the threshold are blocked and weakened.
Applying a high-pass filter to images results in background suppression, edge highlighting, sharpness enhancement, and shape preservation \cite{38}.
A low-pass filter is the opposite of a high-pass filter and can make the image smooth, blurred, and fuzzy.
Usually, for a single channel image $x$, its Fourier transform $\mathcal{F}(x)$ is formulated as:
\begin{equation}
\setlength{\abovedisplayskip}{3pt} 
\setlength{\belowdisplayskip}{3pt}
  \mathcal{F}(x)(u,v)=\sum_{h=0}^{H-1}\sum_{w=0}^{W-1}x(h,w)e^{-j2\pi\left(\frac{h}{H}u+\frac{w}{W}v\right)}.
  \label{eq:0}
\end{equation}
After performing the above Fourier transform, we can express the phase and amplitude components of the image as:
\begin{equation}
\setlength{\abovedisplayskip}{3pt} 
\setlength{\belowdisplayskip}{3pt}
\begin{split}
   \mathcal{P}(x)&{(u,v)}=\arctan\left[\frac{I(x)(u,v)}{R(x)(u,v)}\right],\\
   \mathcal{A}(x)(u,v&{)}={\left[R^2(x)(u,v)+I^2(x)(u,v)\right]}^{\frac{1}{2}},
\end{split}
\label{eq:0.1}
\end{equation}
where $R(x)$ and $I(x)$ represent the real and imaginary part of $\mathcal F(x)$, respectively.
To obtain the corresponding amplitude and phase information, we can compute the Fourier transform independently for each channel of RGB images.

In our method, we use a high-pass filter to filter out low frequency components of the image $x$ by adjusting the diameter $\textit{d}$ on the centered Fourier spectrum:
% \begin{equation}
% \setlength{\abovedisplayskip}{3pt} 
% \setlength{\belowdisplayskip}{3pt}
%   \phi_\textrm{high-pass}{(x)}={H}^d (\mathcal{F}(x))\circ(\mathcal{F}(x)),
%   \label{eq:4}
% \end{equation}
\begin{equation}
\setlength{\abovedisplayskip}{3pt} 
\setlength{\belowdisplayskip}{3pt}
  \phi_\textrm{high-pass}{(x)}=({H}^d\circ\mathcal{F}) (x),
  \label{eq:4}
\end{equation}
where $\circ$ is the function composition symbol, and ${H}^d ({F})$ is a generated high-pass filter mask, respectively.
Each spatial coordinate $(u,v)$ of the mask holds a specific value:
\begin{equation}
\setlength{\abovedisplayskip}{3pt} 
\setlength{\belowdisplayskip}{3pt}
  H_{u, v}^d(F)=\frac{1}{2}\Big[1+\operatorname{sgn}\big(F_{u, v}-d\big)\Big],
  \label{eq:5}
\end{equation}
where $\operatorname{sgn} (\cdot)$ indicates the sign of the parameter.

%\vspace{-0.3cm}
\begin{algorithm}[htbp]
\caption{\small{Two-step High-pass Filter}}
\label{alg:algorithm1}
\LinesNumbered 
\KwIn{Original image $x$, severity level of filter $d$, strength of amplitude scaling $\alpha$, strength of phase scaling $\beta$,  number of images $N$.}
\KwOut{Augmented image $\hat{x}$.}
%some description\; 
\For{$n=1$ to $N$}{
$\mathcal{F}(x)$ $\leftarrow$ Convert $x$ using Eq. (\ref{eq:0})\;
$\phi_\textrm{high-pass}{(x)}$ $\leftarrow$ Filter $\mathcal{F}(x)$ using Eq. (\ref{eq:4}) and Eq. (\ref{eq:5}) with $d$\;
$\mathcal{P}(x)$, $\mathcal{A}(x)$ $\leftarrow$ Obtained using Eq. (\ref{eq:0.1})\;
$\Tilde{\mathcal{A}}(\Tilde{x})$, $\Tilde{\mathcal{P}}(\Tilde{x})$ $\leftarrow$ Scaled using Eq. (\ref{eq:scaling}) with $\alpha$, $\beta$\;
$\mathcal{F}(\hat{x})$ $\leftarrow$ Obtained using Eq. (\ref{eq:7})\;
$\hat{x}$ $\leftarrow$ Transform $\mathcal{F}(\hat{x})$ by $\mathcal{F}^{-1}[\cdot]$ using Eq. (\ref{eq:8}).
}
\end{algorithm}

\textbf{The Second Step: Scaling of Amplitude and Phase.}
In \refsec{sec:gaus}, we mentioned that noise is also a high frequency component.
Therefore, the noise is retained along with the shape of the object in high-pass filtered images. 
Inspired by the fact \cite{69} that an amplitude scaling operation can model the noisy channel, we scale the amplitude and phase of the high-pass filtered Fourier spectrum, respectively:
\begin{equation}
\setlength{\abovedisplayskip}{3pt} 
\setlength{\belowdisplayskip}{3pt}
    \Tilde{\mathcal{A}}(\Tilde{x})=\alpha{\mathcal{A}}(\Tilde{x}),\quad \Tilde{\mathcal{P}}(\Tilde{x})=\beta{\mathcal{P}}(\Tilde{x}),
  \label{eq:scaling}
\end{equation}
where $\alpha, \beta \in \left(0,1\right]$ control the strength of the scaling, and $\Tilde{x}=\phi_\textrm{high-pass}{(x)}$.
In our method, scaling the amplitude and phase of the Fourier transform serves as a technique for mitigating noise within the high-frequency components. 
The objective behind scaling both amplitude and phase is to diminish the intensity of high-frequency elements, consequently minimizing noise. 
(1) Decreasing the amplitude of the Fourier transform effectively diminishes high-frequency components, thereby mitigating their impact. 
Since noise typically manifests as high-frequency vibrations, reducing the amplitude of these high-frequency components can, to some extent, alleviate the contribution of noise.
(2) Scaling the phase alters the distribution of the signal in the frequency domain. 
The phase scaling influences the details and contours of the image, diminishing the impact of noise.
As a consequence of these amplitude and phase scalings, the method helps in diminishing the impact of noise.
Then, the scaled amplitude and phase are combined to form a new Fourier representation, via:
\begin{equation}
\setlength{\abovedisplayskip}{3pt} 
\setlength{\belowdisplayskip}{3pt}
    \mathcal{F}(\hat{x})=\Tilde{\mathcal{A}}(\Tilde{x})
    *e^{-j*{\Tilde{\mathcal{P}}(\Tilde{x})}}.
  \label{eq:7}
\end{equation}
Finally, the inverse FFT  $\mathcal{F}^{-1}[\cdot]$ is applied to transform the new Fourier spectrum $\mathcal{F}(\hat{x})$ back to the spatial domain:
\begin{equation}
\setlength{\abovedisplayskip}{3pt} 
\setlength{\belowdisplayskip}{3pt}
\hat{x}\gets{\mathcal{F}}^{-1}[\mathcal{F}(\hat{x})].
  \label{eq:8}
\end{equation}

Following the above procedure, we generate the Two-step High-pass Filtered augmented image $\hat{x}$ with less noise. 
Algorithm \ref{alg:algorithm1} illustrates the overall process.

\textit{Remark:} Both the frequency restriction and the semantic information extraction strategy employed here differ fundamentally from those used in \cite{5} (FACT) and \cite{107} (BrAD).
(1) In \cite{5}, a direct swap or linear interpolation between the amplitude spectrums of two images from arbitrary source domains is performed to generate augmented images. 
However, this process results in image distortion or pseudo-features, compromising the performance of the model. 
For instance, in the direct swap of amplitude spectrums, the swapped area must be carefully controlled to avoid being too large (which is too aggressive for the model to learn) or too small (which still causes the model to overfit on the remaining amplitude information).
(2) In \cite{107}, the edge model, which could be heuristic, such as Canny edge detector, or pre-trained, such as HED, is used to extract image edges.
This method may confuse or even lose semantic information since the edge model can only capture significant gray value changes and may overlook areas with subtle but semantically meaningful variations.

In contrast, our method focuses on processing images in the frequency domain, allowing us to accurately remove low frequency components and noise while retaining high frequency semantic information, which effectively extracts essential features relevant to the generalization task.
When compared to the above two methods, our method offers several advantages, making it the preferred choice:

\begin{itemize}
\item \textit{Preservation of key semantic information.}
Our method applies the Gaussian Kernel or the high-pass filter in the frequency domain to remove low frequency information and employs amplitude and phase scaling techniques to eliminate noise while retaining high frequency information without noise. 
This method accurately preserves crucial semantic information related to the generalization task without introducing additional pseudo-features. 
However, FACT and BrAD may cause the quality of generated images to degrade due to the introduction of pseudo-features or the retention of noise.

\item \textit{Maintain the integrity and correctness of image generalizable features.}
Our method ensures the integrity and correctness of generalizable features in generated images.
In contrast, the amplitude spectrum swapping or linear interpolation in \cite{5} can lead to image distortion or introduce unrealistic features, while the reliance on edge models in \cite{107} may overlook other important semantic information or introduce irrelevant background edges, reducing the expressiveness of generated images. 

\item \textit{Higher accuracy and stability.}
Our method, rooted in a deep understanding and analysis of image frequency domain characteristics, combines the benefits of filtering and amplitude-phase operations. 
It accurately and completely extracts semantic information while effectively removing noise and irrelevant low frequency information that can interfere with important semantic features and can better support domain generalization tasks. 
High accuracy and stability make our method robust and well adaptable to different scenarios and tasks in a flexible manner.
\end{itemize}

On the contrary, FACT may be affected by the distribution of images in the training set, and the amplitude spectrums are also limited in the source domains. 
The potential low-level statistics are hard to be exhaustively illustrated and fully augmented due to the limited referred amplitude spectrums, leading to the diversity could not be always guaranteed.
BrAD may also be limited by the capabilities of edge models, such as the inability to detect weak edges or the introduction of irrelevant background edges, resulting in decreased generalization ability. 
We experimentally validate the superiority and adaptability of our method in domain generalization tasks.

\vspace{-6pt}
\subsection{Tail Interaction}
\label{sec:tail}
% As mentioned in \cite{63}, self-attention concentrates on the self-affinities between different locations within a single sample but ignores potential correlations with other samples. 
% However, this correlation is meaningful in computer vision. 
In addition to helping models get rid of spurious correlations, we seek to strengthen the potential correlations among samples sourced from various domains, aiming to identify and magnify concealed patterns that hold informative insights into the target task. These patterns encapsulate latent interdependencies and correlations spanning diverse domains. The module facilitates the model in thoroughly investigating and augmenting the exploitation of shared features among samples originating from varied domains. This, in turn, enhances the model's capacity to generalize effectively across different domains.
By incorporating correlations among different samples from various domains, our method aims to learn discriminative representations that are applicable across multiple domains and contain common information. 
Specifically, our method ensures that features belonging to the same category but distributed across different samples are treated consistently.
Consequently, we introduce the Tail Interaction module using two cascaded linear layers and two normalization layers at the end of the network.
These two interaction units implicitly learn the domain-invariant representations across multiple domains.

Given a feature map $F \in \mathbb{R}^{N \times z}$, where $N$ is the number of elements (or pixels in images), and $z$ is the number of feature dimensions, we compute attention between the input pixels and an interaction unit $I_k \in \mathbb{R}^{S \times z}$, via:
\begin{equation}
\setlength{\abovedisplayskip}{3pt} 
\setlength{\belowdisplayskip}{3pt}
A=\left(\alpha\right)_{i,j}=\textrm{DualNorm}\left(FI_k^\top\right),
\label{eq:6}
\end{equation}
where $\alpha_{i,j}$ is the similarity between the $i$-th pixel and the $j$-th row of the interaction unit $I_k$, and $A$ is the normalized attention map. %inferred from this learned dataset-level prior knowledge.
$S$ is a hyper-parameter whose value is set to 64 in our experiments.
The Tail Interaction module allows for manual adjustment of its computational complexity to a much smaller value.
The linear complexity in the number of pixels makes it suitable for high-resolution scenarios. 
We formulate the dual normalization steps as follows:
\begin{equation}
\setlength{\abovedisplayskip}{3pt} 
\setlength{\belowdisplayskip}{3pt}
\begin{split}
&\left(\Tilde\alpha \right)_{i,j}=FI^\top_k,\\
&\hat{\alpha}_{i,j}=\text{exp}\left(\Tilde\alpha _{i,j}\right)/\sum_k\text{exp}\left(\Tilde\alpha _{k,j}\right),\\
&{\alpha}_{i,j}=\hat{\alpha}_{i,j}/\sum_k\hat{\alpha}_{i,k}.
\end{split}
\label{eq:9}
\end{equation}
Then the input features are updated from $I_v \in \mathbb{R}^{S \times z} $ by the similarities in the above normalized attention map $A$, via:
\begin{equation}
\setlength{\abovedisplayskip}{3pt} 
\setlength{\belowdisplayskip}{3pt}
F_{out}=AI_v.
\label{eq:13}
\end{equation}
The two small units, namely $I_k$ and $I_v$, function as learnable parameters independent of the input and work together as crucial interactive roles in acquiring domain-invariant representations across multiple domains. 
The back-propagation algorithm can optimize this module in an end-to-end way instead of requiring an iterative algorithm.

We can further elucidate the mechanism of our module by contrasting it with self-attention. 
In contrast to self-attention, which computes attention on a single sample, our module possesses the ability to establish potential correlations between different samples and acquire domain-invariant representations from features across all samples and domains.

\begin{figure}[h]
  \centering
\includegraphics[width=1\linewidth]{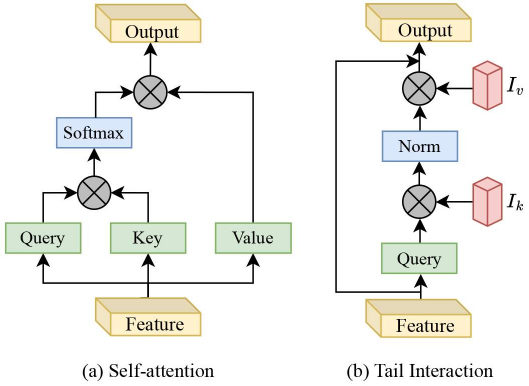}
   \caption{Diagrams of Self-attention and Tail Interaction.}
   \label{fig:tail}
\end{figure}

Self-attention mechanisms play a crucial role in computer vision by allowing models to focus selectively on the most critical parts of an image. 
Self-attention involves self-learning and weight distribution within a feature map, creating an attention mechanism between internal elements. 
As shown in \reffig{fig:tail}, to compute self-attention, we first calculate an attention map by computing the affinities between self-query vectors and self-key vectors. 
Then, we generate a new feature map by weighting the self-value vectors with this attention map.

In contrast, Tail Interaction first calculates the attention map by computing the affinities between the self-query vectors and a \textit{key} learnable interaction unit and then produces a refined reconstructed feature map by multiplying this attention map by another learnable \textit{value} interaction unit. 
The processes involved can be viewed as projecting the self-query of all samples into a shared space instead of onto the self-key and then reconstructing the projection with another interactive unit back to features.
All samples from all domains constrain this shared space, which plays a dual role of regularization and generalization in learning domain-invariant representations, implicitly considering the correlations among all data samples from various domains.
Moreover, the interaction unit (shared space) is highly flexible, which allows us to control its size and make the module adaptable for high-resolution applications.

Typically, the self-attention mechanism employs a single softmax as the central normalization layer.
This normalization results in each row or column of the attention map summing up to 1, \eg, $\sum_j \alpha_{i,j}=1$.
However, using only softmax can cause a feature value to obtain a relatively large or small dot product with other features when the value itself is relatively large or small.
In this case, using only softmax destroys the original meaning of the self-attention mechanism. 
To overcome this issue, researchers have introduced the double normalization method \cite{63,89}, which uses softmax on the first dimension followed by $l_1  \textrm{-norm}$ on the second dimension to normalize the attention map.
Consequently, the Tail interaction module can be implemented using only two linear layers without bias (realizing the matrix multiplication) and two normalization layers, as shown in Algorithm \ref{alg:algorithm2}.

\begin{algorithm}[htbp]
\caption{Tail Interaction}
\label{alg:algorithm2}
\LinesNumbered 
\KwIn{The feature map $F \in \mathbb{R}^{B \times N \times C}$ (batch size $B$, number of pixels $N$, number of channels $C$); query linear $Q_l$; two linear layers $I_k$ and $I_v$; softmax operator $N_{s}$; $l_1  \textrm{-norm}$ operator $N_{l_1}$.}
\KwOut{The refined feature map $F_{out} \in \mathbb{R}^{B \times N \times C}$.}

\hspace{0.5em}$\hat F$ $\leftarrow$ $F$ \tcp*[l]{{\scriptsize \textcolor{gray}{Deep clone for residual connection.}}}
\hspace{0.5em}$\hat F$ $\leftarrow$ Obtained by $Q_l$($\hat F$) \tcp*[l]{{\scriptsize \textcolor{gray}{Self-query vector.}}}
\hspace{0.5em}$F_{atten}$ $\leftarrow$ Computed by $I_k$ ($\hat F$)  \tcp*[l]{{\scriptsize \textcolor{gray}{Linear layer}.}}
\hspace{0.5em}$F_{atten1}$ $\leftarrow$ $N_{s}$($F_{atten}, dim=1$)  \tcp*[l]{{\scriptsize \textcolor{gray}{Softmax.}}}
% normalized by softmax on the first dimension\;
\hspace{0.5em}$F_{atten2}$ $\leftarrow$ $N_{l_1}$($F_{atten1}, dim=2$) \tcp*[l]{{\scriptsize \textcolor{gray}{$l_1$-norm.}}}
% attention $\leftarrow$ normalized by $l_1 \textrm{-norm}$ on the second dimension\;
\hspace{0.5em}$\hat F$ $\leftarrow$ Computed by $I_v (F_{atten2})$\tcp*[l]{{\scriptsize \textcolor{gray}{Linear layer.}}}
% out $\leftarrow$ obtained by $I_v (\textrm{attention})$\;
\hspace{0.5em}$\hat F$ $\leftarrow$ $\hat F$ + $F$ \tcp*[l]{{\scriptsize \textcolor{gray}{Residual connection.}}}
\hspace{0.5em}$F_{out}$ $\leftarrow$ $ReLU(\hat F)$ \tcp*[l]{{\scriptsize \textcolor{gray}{ReLU activation.}}}
% obtained by using activation function on out
\end{algorithm}

\textit{Remark:} We notice that Batchformer \cite{75} also explores sample relationships during training.
Nevertheless, we are entirely different.
The BatchFormer module employs a stack of transformer encoder layers, which include multi-head self-attention, to model the relationships between different samples in each mini-batch.
Our Tail Interaction module is distinguished from the BatchFormer by its $O(n)$ linear complexity and implicit consideration of correlations among all samples, in contrast to the quadratic complexity of self-attention and the relationship establishment between samples in a mini-batch.
This difference is significant and sets us apart.

%------------------------------------------------------------------------
%\vspace{-5pt}
\section{Experiments}
\label{sec:experiments}
We organize the section of experiments as follows.
(1) Description of the four widely-used datasets in DG tasks: \refsec{sec:datasets}.
(2) Implementation details of experiments: \refsec{sec:detail} and the supplementary material.
(3) Baseline models for comparison: \refsec{sec:baseline}.
(4) Results of embedding our modules into CNNs and MLPs: \refsec{sec:result} and the supplementary material.
(5) Ablation studies: \refsec{sec:ablation} and the supplementary material.
(6) Further analysis: \refsec{sec:analysis} and the supplementary material.
(7) Results on DomainBed protocol: the supplementary material.

\subsection{Datasets}
\label{sec:datasets}
\textbf{PACS} contains 9,991 images with 7 classes and 4 domains, \ie, Photo, Art painting, Cartoon, and Sketch. 
PACS is regarded as the most commonly used DG dataset, which includes significant distribution discrepancies across domains. 
We apply the official train-validation split to conduct the experiments for a fair comparison.

\textbf{VLCS} consists of 10,729 images, including 5 classes and 4 domains, \ie, Caltech-101, LabelME, PASCAL VOC 2007, and Sun09.
%It is also a popular DG dataset.
We adopt the official split of VLCS in our experiments for a fair comparison.

\textbf{Office-Home} contains around 15,500 images, covering 65 categories and 4 domains \ie, Artistic, Clipart, Product, and Real-World. 
Its number of classes is significantly larger than that of PACS and VLCS, and it is challenging in DG tasks.
For a fair comparison, we randomly split the whole dataset into 90$\%$:10$\%$ as training and validation sets following \cite{17}.

\textbf{Digits-DG} is a combination of four handwritten digit recognition datasets, \ie, MNIST, MNIST-M, SVHN, and SYN, with domain shift mainly in font style, stroke color, and background.
Following \cite{45}, we randomly select 600 images per class for each domain and split 80$\%$ of the data as training and 20$\%$ as validation.

\subsection{Implementation Details}
\label{sec:detail}
For the experiments on CNNs, the implementation details are as follows.
% For PACS, VLCS, and OfficeHome, we use ImageNet \cite{49} pre-trained AlexNet \cite{79}, ResNet-18, and ResNet-50 \cite{32} as the backbones.
For PACS, VLCS, and OfficeHome, we use ImageNet pre-trained AlexNet, ResNet-18, and ResNet-50 as the backbones.
All images are resized to $224 \times 224$.
For Digits-DG, we use the same backbone network as \cite{14,45}, and the input image size is $32 \times 32$. 
The batch size is set to 64. 
The typical data augmentation consists of random resize and crop with an interval of [0.8, 1], random horizontal flip, and random color jittering with a ratio of 0.4. 
We train our model using the Nesterov-momentum SGD with a momentum of 0.9 and weight decay of 5e-4. 
For PACS, VLCS, and Digits-DG, we train the model for 50 epochs. 
For Office-Home, we train the model for 30 epochs.  
The initial learning rate for PACS, VLCS, and OfficeHome is 0.001, decayed by 0.1 at 80$\%$ of the total epochs. 
The initial learning rate for Digits-DG is 0.05, decayed by 0.1 every 20 epochs.
We employ Gaussian Kernel with a kernel size of 63.
We randomly select the severity level $d$ of the high-pass filter for each original image from five levels.
The scaling ratio $\alpha$ and $\beta$ are randomly selected in the interval $[0.6,0.7,0.8,0.9,1.0]$.
We pass the original images, as well as their augmented counterparts, to the model.
Therefore, within a single iteration, the number of input images is $2\times$batch size.
We adopt the ``leave-one-domain-out" \cite{41} experimental protocol that leaves one domain as an unseen domain and others as source domains.
Following the way in \cite{41}, we select the best model on the validation set.
We report top-1 classification accuracy ($\%$) averaged over five runs with different seeds for performance measures.

For taking MLPs as backbones, we choose GFNet-H-Ti (15M) and GFNet-H-S (32M) \cite{30} pre-trained on ImageNet, which maintain comparable parameters with ResNet-18 (12M) and ResNet-50 (26M).
With the DG modules we use (only Tail Interaction add parameters), the network increases the number of parameters by no more than 65,000 and remains comparable.
The number of blocks in the four stages of the GFNet-H-Ti model and the GFNet-H-S model is $[3, 3, 10, 3]$.
The number of channels in the four stages is $[64, 128, 256, 512]$ for GFNet-H-Ti and $[96, 192, 384, 768]$ for GFNet-H-S, respectively.
We utilize 4$\times$4 patch embedding to form the input tokens and a non-overlapping convolution layer to downsample the tokens following \cite{30}. 
All images are resized to 224$\times$224, and the batch size is set to 64 for training.
The standard data augmentation \cite{17} consists of random resize and crop with an interval of [0.8, 1], random horizontal flipping, and random color jittering with a ratio of 0.4. 
We train our model for 30 epochs using the AdamW \cite{50} optimizer and set the weight decay to 0.05. 
We set the initial learning rate as 5e-4 with cosine decay.
Linear learning rate warm-up in the first five epochs and gradient clipping with a max norm of 1 are also adopted to stabilize the training process.

\begin{table*}[t]

\large
\centering
\caption{DG accuracy ($\%$) on PACS.
The best performance is marked as \textbf{bold} for different backbones, respectively.
A, C, P, and S indicate the target domain of Art painting, Cartoon, Photo, and Sketch.
G in brackets means using Gaussian Kernel, and T means using Two-step High-pass Filter.
The backbones of MLPs are GFNet-H-Ti and GFNet-H-S
}
\resizebox{\linewidth}{!}{
\begin{tabular}{l|c|cccc>{\columncolor{gray!15}}c|cccc>{\columncolor{gray!15}}c}
\hline
\toprule[0.5pt]
\specialrule{0em}{1pt}{1pt}
\textbf{Methods}               & \textbf{Venue}                & \textbf{A} & \textbf{C} & \textbf{P} & \textbf{S} & \textbf{Avg.} & \textbf{A} & \textbf{C} & \textbf{P} & \textbf{S} & \textbf{Avg.} \\ \specialrule{0em}{1pt}{1pt}\hline \specialrule{0em}{1pt}{1pt}
\multicolumn{1}{c|}{\textit{}} & \multicolumn{1}{l|}{}         & \multicolumn{5}{c|}{\textit{ResNet-18}}                            & \multicolumn{5}{c}{\textit{ResNet-50}}                            \\ \specialrule{0em}{1pt}{1pt} \hline
\specialrule{0em}{1pt}{1pt}
% ER \cite{109}                      & NeurIPS'20                       & -      & -      & -      & -      & -         & 87.50      & 79.30      & 98.30     & 76.30      & 85.30         \\
RSC \cite{56}                      & ECCV'20                       & 83.43      & 80.31      & 95.99      & 80.85      & 85.15         & 87.89      & 82.16      & 97.92      & 83.85      & 87.83         \\
L2A-OT \cite{4}                      & ECCV'20                       & 83.30      & 78.20      & 96.20      & 73.60      & 82.80         & -      & -      & -     & -      & -         \\
DeepAll \cite{5}                       & CVPR'21                     & 77.63      & 76.77      & 95.85      & 69.50      & 79.94         & 84.94      & 76.98      & 97.64      & 76.75      & 84.08         \\
SFA \cite{81}                    & ICCV'21                       & 80.10      & 76.20      & 94.10      & 73.50      & 81.00        & -      & -      & -      & -      & -         \\
COPA \cite{82}                    & ICCV'21                       & 83.30      & 79.80      & 94.60      & 82.50      & 85.10        & -      & -      & -      & -      & -         \\
STEAM \cite{80}                    & ICCV'21                       & 85.50      & 80.60      & \textbf{97.50}      & 82.90      & 86.60         & -      & -      & -      & -      & -         \\
FACT \cite{5}                    & CVPR'21                       & 85.37      & 78.38      & 95.15      & 79.15      & 84.51         & 89.63      & 81.77      & 96.75      & 84.46      & 88.15         \\
I$^2$-ADR \cite{108}                    & ECCV'22                       & 82.90      & 80.80      & 95.00      & 83.50      & 85.60         & 88.50      & 83.20      & 95.20      & 85.80      & 88.20         \\
MVDG \cite{55}                    & ECCV'22                       & 85.62      & 79.98      & 95.54      & 85.08      & 86.56         & 89.31      & 84.22      & 97.43      & 86.36      & 89.33         \\
StyleNeophile \cite{51}          & CVPR'22                       & 84.41      & 79.25      & 94.93      & 83.27      & 85.47         & 90.35      & 84.20      & 96.73      & 85.18      & 89.11         \\
GeomText \cite{52}                & CVPR'22                       & \textbf{86.99}      & 80.38      & 96.68      & 82.18      & 86.56         & 89.98      & 83.84      & 98.10      & 84.75      & 89.17         \\
CIRL \cite{54}                    & CVPR'22                       & 86.08      & 80.59      & 95.93      & 82.67      & 86.32         & \textbf{90.67}      & 84.30      & 97.84      & \textbf{87.68}      & 90.12         \\
EFDMix \cite{74}                    & CVPR'22                       & 83.90      & 79.40      & 96.80      & 75.00      & 83.90         & 90.60      & 82.50      & 98.10      & 76.40      & 86.90         \\
PCL \cite{10}                    & CVPR'22                       & -      & -      & -      & -      & -         & 90.20      & 83.90      & 98.10      & 82.60      & 88.70         \\
P-RC \cite{88}                    & CVPR'23                       & 83.15      & 81.07      & 96.24      & 76.71      & 84.29         & 89.28      & 84.13      & 97.83      & 81.85      & 88.27         \\
\textbf{Ours (G)}                    & This paper                       & 83.73\tiny$\pm$0.48      & \textbf{81.40}\tiny$\pm$0.92      & 96.89\tiny$\pm$0.33      & 85.07\tiny$\pm$0.71      & 86.77         & 90.25\tiny$\pm$0.50      & 86.17\tiny$\pm$0.46     & 98.52\tiny$\pm$0.25     & 85.94\tiny$\pm$0.49      & \textbf{90.22}         \\
\textbf{Ours (T)}                   & This paper                       & 83.51\tiny$\pm$0.65      & 81.34\tiny$\pm$1.09      & 97.20\tiny$\pm$0.41      & \textbf{85.11}\tiny$\pm$0.90      & \textbf{86.79}         & 90.11\tiny$\pm$0.47      & \textbf{86.22}\tiny$\pm$0.88     & \textbf{98.98}\tiny$\pm$0.26     & 85.57\tiny$\pm$0.76      & \textbf{90.22}         \\
 \specialrule{0em}{1pt}{1pt}\hline\specialrule{0em}{1pt}{1pt}
                               & \multicolumn{1}{l|}{}         & \multicolumn{10}{c}{\textit{MLPs}}                                                                \\ \specialrule{0em}{1pt}{1pt}\hline\specialrule{0em}{1pt}{1pt}
% DoPrompt \cite{21}                 & \multicolumn{1}{c|}{Arxiv'22} & -          & -          & -          & -          & -             & 91.10      & 83.00      & \textbf{99.60}      & 78.70      & 88.10         \\
% SDViT \cite{58}                   & \multicolumn{1}{c|}{Arxiv'22} & -          & -          & -          & -          & -             & 91.20      & 83.50      & 98.30      & 82.50      & 88.90         \\ 
FAMLP \cite{20}                   & \multicolumn{1}{c|}{Arxiv'22} & \textbf{92.06}      & 82.49      & 98.10      & 84.09      & 89.19         & 92.63      & \textbf{87.03}      & 98.14      & 82.69      & 90.12         \\
GFNet Baseline                    & \multicolumn{1}{c|}{This paper} & 91.20\tiny$\pm$0.85      & 82.28\tiny$\pm$1.45      & \textbf{99.03}\tiny$\pm$0.14      & 78.48\tiny$\pm$1.28      & 87.75         & 91.42\tiny$\pm$0.14      & 84.88\tiny$\pm$0.50     & 98.23\tiny$\pm$0.18      &79.95\tiny$\pm$0.32      &  88.62       \\
\textbf{Ours (G)}                           & This paper                    & 91.05\tiny$\pm$0.14      & \textbf{83.23}\tiny$\pm$1.02      & 98.67\tiny$\pm$0.01      & 85.01\tiny$\pm$0.47      & 89.49         & 92.68\tiny$\pm$0.23      & 86.22\tiny$\pm$0.59      & 99.04\tiny$\pm$0.08      & 85.76\tiny$\pm$1.26      & 90.93         \\
\textbf{Ours (T)}                           & This paper                    & 91.07\tiny$\pm$0.15      & 83.17\tiny$\pm$1.71      & 98.76\tiny$\pm$0.05      & \textbf{85.19}\tiny$\pm$1.06      & \textbf{89.55}         & \textbf{92.72}\tiny$\pm$0.18      & 86.46\tiny$\pm$0.61      & 98.82\tiny$\pm$0.02      & \textbf{85.79}\tiny$\pm$1.33      & \textbf{90.95}         \\ \specialrule{0em}{1pt}{1pt}\hline
\toprule[0.5pt]
\end{tabular}}
%\vspace{-15pt}
\label{tab:PACS}
\end{table*}

\subsection{Baselines}
\label{sec:baseline}
We introduce conventional DeepAll as a baseline, \ie, the pre-trained backbones fine-tuned on the aggregation of all source domains without applying any other domain generalization modules, with only classification loss.

We also raise several representative methods from the categories described in \refsec{sec:ralatedWork} as baseline methods: \textit{domain-invariant representations learning based} methods: FACT \cite{5}, StyleNeophile \cite{51}, GeomText \cite{52}, MVDG \cite{55}, and COMEN \cite{60}, \textit{adversarial learning based} method: MMLD \cite{61}, \textit{gradient/neuron diversity based} method: RSC \cite{56}, and \textit{samples weights based} method: StableNet \cite{62}. 
Besides, we compare with recently proposed state-of-the-art methods: SFA \cite{81}, COPA \cite{82}, STEAM \cite{80}, CIRL \cite{54}, EFDMix \cite{74}, P-RC \cite{88}, GINet \cite{106}, and MLPs based FAMLP \cite{20}.

Please note that the baseline models listed in the result tables for different datasets are not identical. 
The related and the most advanced works use different datasets (\eg, VLCS is not used in \cite{5}) or inconsistent backbones for the same dataset.
We do not deliberately omit the results of the model used for comparison on any dataset.

\begin{table}[t]
  \setlength{\belowcaptionskip}{-0.7cm}
   \setlength{\abovecaptionskip}{-0.1cm}
\centering
\Large
\caption{DG accuracy ($\%$) on VLCS.
The best performance is marked as \textbf{bold} for different backbones, respectively.
C, L, P, and S indicate the target domain of Caltech-101, LabelMe, PASCAL VOC2007, and Sun09.
G in brackets means using Gaussian Kernel, and T means using Two-step High-pass Filter.
The backbone of MLPs is GFNet-H-Ti
}
\resizebox{\linewidth}{!}{
\begin{tabular}{l|c|cccc>{\columncolor{gray!15}}c}
\specialrule{0em}{1pt}{1pt}\hline\toprule[0.5pt]\specialrule{0em}{1pt}{1pt}
\textbf{Methods}               & \textbf{Venue}        & \textbf{C} & \textbf{L} & \textbf{P} & \textbf{S} & \textbf{Avg.} \\ \specialrule{0em}{1pt}{1pt}\hline\specialrule{0em}{1pt}{1pt}
\multicolumn{1}{c|}{\textit{}} & \multicolumn{1}{l|}{} & \multicolumn{5}{c}{\textit{AlexNet}}                              \\ \specialrule{0em}{1pt}{1pt}\hline\specialrule{0em}{1pt}{1pt}
DeepAll \cite{56}                       &ECCV'20             & 96.25      & 59.72      & 70.58      & 64.51      & 72.76         \\
RSC \cite{56}                     & ECCV'20               & 97.61      & 61.86      & 73.93      & 68.32      & 75.43         \\
MMLD \cite{61}                    & AAAI'20               & 96.66      & 58.77      & 71.96      & 68.13      & 73.88         \\
SFA \cite{81}                   & ICCV'21               & 96.20      & 62.20      & 69.60      & 65.80      & 73.50         \\
COMEN \cite{60}                   & CVPR'22               & 97.00      & 62.60      & 72.80      & 67.60      & 75.00         \\
P-RC \cite{88}                   & CVPR'23               & -      & -      & -      & -      & 71.30         \\
\textbf{Ours (G)}                   & This paper               & 97.58\tiny$\pm$0.35      & \textbf{62.75}\tiny$\pm$0.81     & 74.39\tiny$\pm$0.54     & 69.24\tiny$\pm$0.92      & 75.99        \\
\textbf{Ours (T)}                   & This paper               & \textbf{97.64}\tiny$\pm$0.44      & \textbf{62.75}\tiny$\pm$0.50     & \textbf{74.51}\tiny$\pm$0.75     & \textbf{69.38}\tiny$\pm$0.76      & \textbf{76.07}        \\
\specialrule{0em}{1pt}{1pt}\hline\specialrule{0em}{1pt}{1pt}
\multicolumn{1}{c|}{\textit{}} &                       & \multicolumn{5}{c}{\textit{ResNet-18}}                                 \\ \specialrule{0em}{1pt}{1pt}\hline\specialrule{0em}{1pt}{1pt}
DeepAll  \cite{62}                          & CVPR'21            & 91.86  & 61.81     & 67.48     & 68.77     & 72.48        \\
StableNet \cite{62}                          & CVPR'21            & 96.67   &\textbf{65.36}     & 73.59      & 74.97     & 77.65        \\
MVDG \cite{55}                          & ECCV'22           & \textbf{98.40}     & 63.79      & 75.26     & 71.05     &  77.13      \\
P-RC \cite{88}                   & CVPR'23               & -      & -      & -      & -      & 69.55         \\
\textbf{Ours (G)}                   & This paper               & 98.03\tiny$\pm$0.41      & 63.88\tiny$\pm$0.55     & 74.27\tiny$\pm$0.91     & 74.99\tiny$\pm$1.02      & 77.79        \\
\textbf{Ours (T)}                          & This paper             & 97.92\tiny$\pm$0.56      & 63.65\tiny$\pm$0.21     & \textbf{75.84}\tiny$\pm$0.80      & \textbf{75.03}\tiny$\pm$0.66      & \textbf{78.11}      \\
 \specialrule{0em}{1pt}{1pt}\hline\specialrule{0em}{1pt}{1pt}
\multicolumn{1}{c|}{\textit{}} &                       & \multicolumn{5}{c}{\textit{MLPs}}                                 \\ \specialrule{0em}{1pt}{1pt}\hline\specialrule{0em}{1pt}{1pt}
GFNet Baseline                           & This paper            & \textbf{98.79}\tiny$\pm$0.12      & 62.70\tiny$\pm$0.13      & 78.09\tiny$\pm$0.34      & 74.80\tiny$\pm$0.18      &78.60          \\
\textbf{Ours (G)}                          & This paper            & 98.45\tiny$\pm$0.15      & \textbf{65.44}\tiny$\pm$1.67      & \textbf{81.00}\tiny$\pm$1.35      & 75.65\tiny$\pm$1.47      & 80.13         \\
\textbf{Ours (T)}                           & This paper            & 98.49\tiny$\pm$0.28      & 65.04\tiny$\pm$0.83      & \textbf{81.00}\tiny$\pm$1.06      & \textbf{76.31}\tiny$\pm$0.96      & \textbf{80.21}         \\
\specialrule{0em}{1pt}{1pt}\hline\toprule[0.5pt]\specialrule{0em}{1pt}{1pt}
\end{tabular}}
\label{tab:VLCS}
\vspace{-15pt} 
\end{table}

\begin{table}[htbp]
\setlength{\belowcaptionskip}{-0.5cm}
\centering
\Large
\caption{DG accuracy ($\%$) on Office-Home.
The best performance is marked as \textbf{bold} for different backbones, respectively.
A, C, P, and R indicate the target domain of Artistic, Clipart, Product, and Real-World.
G in brackets means using Gaussian Kernel, and T means using Two-step High-pass Filter.
The backbone of MLPs is GFNet-H-Ti
}
\resizebox{\linewidth}{!}{
\begin{tabular}{l|c|cccc>{\columncolor{gray!15}}c}
\specialrule{0em}{1pt}{1pt}\hline\toprule[0.5pt]\specialrule{0em}{1pt}{1pt}
\textbf{Methods}               & \textbf{Venue}        & \textbf{A} & \textbf{C} & \textbf{P} & \textbf{R} & \textbf{Avg.} \\ \specialrule{0em}{1pt}{1pt}\hline\specialrule{0em}{1pt}{1pt}
\multicolumn{1}{c|}{\textit{}} & \multicolumn{1}{l|}{} & \multicolumn{5}{c}{\textit{ResNet-18}}                            \\ \specialrule{0em}{1pt}{1pt}\hline\specialrule{0em}{1pt}{1pt}
RSC \cite{56}                    & ECCV'20               & 58.42      & 47.90      & 71.63      & 74.54      & 63.12         \\
L2A-OT \cite{4}                    & ECCV'20               & 60.60      & 50.10      & 74.80      & 77.00      & 65.60         \\
DeepAll \cite{5}                       &CVPR'21            & 57.88      & 52.72      & 73.50      & 74.80      & 64.72         \\
COPA \cite{82}                    & ICCV'21               & 59.40      & 55.10      & 74.80      & 75.00      & 66.10         \\
STEAM \cite{80}                    & ICCV'21               & 62.10      & 52.30      & 75.40      & 77.50      & 66.80         \\
FACT \cite{5}                    & CVPR'21               & 60.34      & 54.85      & 74.48      & 76.55      & 66.56         \\
StyleNeophile \cite{51}           & CVPR'22               & 59.55      & 55.01      & 73.57      & 75.52      & 65.89         \\
GeomText \cite{52}                & CVPR'22               & 60.40      & \textbf{56.30}      & 74.55      & 75.77      & 66.75    \\
CIRL \cite{54}                    & CVPR'22               & 61.48      & 55.28      & 75.06      & 76.64      & 67.12         \\
MVDG \cite{55}                    & ECCV'22               & 60.25      & 54.32      & 75.11      & 77.52      & 66.80         \\ 
P-RC \cite{88}                   & CVPR'23               & -      & -      & -      & -      & 64.59         \\
GINet \cite{106}                   & TIP'23               & 61.90      & 52.70      & 75.30      & 77.50      & 66.90         \\
\textbf{Ours (G)}                   & This paper               & 62.48\tiny$\pm$0.81      & 53.93\tiny$\pm$0.76      & 75.66\tiny$\pm$0.35      &77.51\tiny$\pm$0.69      & 67.40         \\
\textbf{Ours (T)}                   & This paper               & \textbf{62.67}\tiny$\pm$0.94      & 54.04\tiny$\pm$0.41      & \textbf{75.85}\tiny$\pm$0.25      &\textbf{77.80}\tiny$\pm$0.44      & \textbf{67.59}         \\\specialrule{0em}{1pt}{1pt}\hline\specialrule{0em}{1pt}{1pt}
& \multicolumn{1}{l|}{} & \multicolumn{5}{c}{\textit{MLPs}}                                 \\ \specialrule{0em}{1pt}{1pt}\hline\specialrule{0em}{1pt}{1pt}
FAMLP \cite{20}                   & Arxiv'22              & 69.34      & \textbf{62.61}      & 79.82      & \textbf{82.00}      & 73.44         \\
GFNet Baseline                           & This paper            & 66.82\tiny$\pm$0.12      & 55.47\tiny$\pm$0.06     & 78.91\tiny$\pm$0.11      & 80.33\tiny$\pm$0.26      & 70.38         \\
\textbf{Ours (G)}                           & This paper            & 70.70\tiny$\pm$0.09      & 60.92\tiny$\pm$0.45      & 80.00\tiny$\pm$0.05      & 81.52\tiny$\pm$0.06      & 73.29         \\
\textbf{Ours (T)}                           & This paper            & \textbf{71.24}\tiny$\pm$0.13      & 62.21\tiny$\pm$0.35      & \textbf{80.90}\tiny$\pm$0.10      & 81.97\tiny$\pm$0.11      & \textbf{74.08}         \\ \specialrule{0em}{1pt}{1pt}\hline\toprule[0.5pt]\specialrule{0em}{1pt}{1pt}
\end{tabular}}
\label{tab:OH}
\vspace{-10pt} 
\end{table}

\begin{table}[htbp]
%\vspace{3pt} 
\centering
\Large
\caption{DG accuracy ($\%$) on Digits-DG.
The best performance is marked as \textbf{bold} for different backbones, respectively.
MN, MN-M, SV, and SY indicate the target domain of MNIST, MNIST-M, SVHN, and SYN.
G in brackets means using Gaussian Kernel, and T means using Two-step High-pass Filter.
The backbone of MLPs is GFNet-H-Ti
}
\resizebox{\linewidth}{!}{
\begin{tabular}{l|c|cccc>{\columncolor{gray!15}}c}
\specialrule{0em}{1pt}{1pt}\hline\toprule[0.5pt]\specialrule{0em}{1pt}{1pt}
\textbf{Methods}               & \textbf{Venue}        & \textbf{MN} & \textbf{MN-M} & \textbf{SV} & \textbf{SY} & \textbf{Avg.} \\ \specialrule{0em}{1pt}{1pt}\hline\specialrule{0em}{1pt}{1pt}
\multicolumn{1}{c|}{\textit{}} & \multicolumn{1}{l|}{} & \multicolumn{5}{c}{\textit{Backbone in \cite{14,45}}}                                  \\ \specialrule{0em}{1pt}{1pt}\hline\specialrule{0em}{1pt}{1pt}
DeepAll \cite{45}                       &AAAI'20             & 95.8           & 58.8             & 61.7          & 78.6         & 73.7          \\
L2A-OT \cite{4}                    & ECCV'20               &96.7            & 63.9             & 68.6        & 83.2         & 78.1         \\
SFA \cite{81}                    & ICCV'21               &96.7            & 66.3             & 68.8         & 85.1         & 79.2          \\
COPA \cite{82}                    & ICCV'21               &97.0            & 66.5             & 71.6         & 90.7         & 81.5          \\
STEAM \cite{80}                    & ICCV'21               & 96.8           & 67.5             & 76.0          & \textbf{92.2}         & 83.1          \\
FACT \cite{5}                    & CVPR'21               & \textbf{97.9}           & 65.6             & 72.4          & 90.3         & 81.5          \\
GeomText \cite{52}                & CVPR'22               & 97.8           & 70.0             & 75.1          & 87.3         & 82.6          \\

COMEN \cite{60}                   & CVPR'22               & 97.1           & 67.6             & 75.1          & 91.3         & 82.3          \\
CIRL \cite{54}                    & CVPR'22               & 96.1           & 69.9             & 76.2          & 87.7         & 82.5          \\ 
\textbf{Ours (G)}                  & This paper               & 96.1\tiny$\pm$0.1      & 70.1\tiny$\pm$0.3      & 75.9\tiny$\pm$0.2      & 91.5\tiny$\pm$0.4      & 83.4         \\
\textbf{Ours (T)}                  & This paper               & 96.4\tiny$\pm$0.2      & \textbf{70.3}\tiny$\pm$0.3      & \textbf{76.4}\tiny$\pm$0.4      & 91.5\tiny$\pm$0.4      & \textbf{83.7}         \\
 \specialrule{0em}{1pt}{1pt}\hline\specialrule{0em}{1pt}{1pt}
\multicolumn{1}{c|}{\textit{}} &                       & \multicolumn{5}{c}{\textit{MLPs}}                                                \\ \specialrule{0em}{1pt}{1pt}\hline\specialrule{0em}{1pt}{1pt}
FAMLP \cite{20}                   & Arxiv'22              & \textbf{98.0}           & 83.3             & 84.1          & \textbf{96.9}         & 90.6          \\
GFNet Baseline                    & This paper              & \textbf{98.0}\tiny$\pm$0.2           & 74.1\tiny$\pm$0.2             & 80.6\tiny$\pm$0.4          & 96.6\tiny$\pm$0.3         & 87.3          \\
\textbf{Ours (G)}                           & This paper            & \textbf{98.0}\tiny$\pm$0.1           & 86.7\tiny$\pm$0.4             & 85.0\tiny$\pm$0.6          & 96.6\tiny$\pm$0.1         & 91.6          \\
\textbf{Ours (T)}                           & This paper            & \textbf{98.0}\tiny$\pm$0.1           & \textbf{88.9}\tiny$\pm$0.5             & \textbf{85.3}\tiny$\pm$0.6          & \textbf{96.9}\tiny$\pm$0.2         & \textbf{92.3}          \\ \specialrule{0em}{1pt}{1pt}\hline\toprule[0.5pt]\specialrule{0em}{1pt}{1pt}
\end{tabular}
}
\label{tab:digits}
%\vspace{-20pt} 
\end{table}

\vspace{-2pt} 
\subsection{Results}
\label{sec:result}
\textbf{PACS.} 
We summarise our observations from Table \ref{tab:PACS} as follows.
(1) As demonstrated, the models surpass other DG methods when we embed our modules into CNNs or MLPs.
(2) Our models based on MLPs with Two-step High-pass Filter perform best (89.55$\%$ and 90.95$\%$) compared to the others.
The Gaussian Kernel and the Two-step High-pass Filter make the model learn domain-invariant representations, which are relevant features of objects and category labels. 
Thus, our models achieve higher generalization accuracy than others.
(3) The baseline performance of GFNet-H-Ti (87.75$\%$) exceeds that of all listed methods based on ResNet-18, providing a strong baseline for DG.

\textbf{VLCS.} 
In Table \ref{tab:VLCS}, we present the performance of the CNNs and MLPs embedded with our modules on VLCS and their comparison with other methods.
The CNNs with our modules (T) (76.07$\%$ and 78.11$\%$) outperform the state-of-the-art, indicating the usefulness of shape information in this setting. 
Besides, the GFNet-H-Ti baseline result (78.60$\%$) has surpassed all DG methods based on CNNs.
Our results with MLPs backbone further confirm the effectiveness of our two frequency restriction strategy and the Tail Interaction module, building a new state-of-the-art.

\textbf{Office-Home.} 
We observe from Table \ref{tab:OH} that DeepAll has achieved impressive performance (64.72$\%$) because the dataset exhibits a smaller domain gap compared to PACS, especially in the Artistic, Product, and Real-World domains, where variations primarily exist in the background and viewpoint.
Among all methods, our model based on CNNs achieves a clear margin (67.59$\%$) against other methods.
Moreover, the model performs better when embedding our modules into MLPs with a strong baseline than the most advanced methods. 
The results powerfully demonstrate the versatility of our models.

\textbf{Digits-DG.} 
As observed in Table \ref{tab:digits}, our model (T) based on CNNs or GFNet-H-Ti achieves the best performance.
On the two most challenging target domains, MNIST-M and SVHN, which contain complex backgrounds and cluttered digits, respectively, our model (T) based on MLPs obtains clear margins over the competitors, notably with +5.6$\%$ and +1.2$\%$ improvements compared with the FAMLP model. 
Results show that our models can effectively deal with significant domain shifts caused by complex backgrounds and cluttered digits.
The baseline results of GFNet-H-Ti (87.3$\%$) exceed the best performance of the convolution network-based method, as is the case on other datasets.

\subsection{Ablation Studies}
\label{sec:ablation}
We conduct extensive ablation studies to investigate the role of each module in our framework, as shown in Table \ref{tab:ab0}. 
Model A is trained solely using the High-Pass (HP) filter data augmentation from the baseline. 
Model B is derived from Model A by incorporating the Tail Interaction (TI) module, which leads to better performance. 
We train Models C and D using only HP, TI, and the other scaling to investigate the necessity of scaling of phase or amplitude.
However, neither model C nor D outperforms the complete network, indicating the significance of incorporating both scalings to reduce noise during domain-invariant representations learning. 
Our complete network, which includes Phase Scaling (PS) and Amplitude Scaling (AS), outperforms all other models, highlighting the importance of these scalings in reducing noise interference and model generalization.
Furthermore, excluding the Tail Interaction to create Model E results in decreased performance, indicating its indispensable role.
In fact, Model E achieves an accuracy rate of 85.08$\%$ using only the data augmentation strategy, surpassing FACT \cite{5} (84.51$\%$) as shown in Table \ref{tab:PACS}. 
It is worth noting that FACT combines both Fourier-based data augmentation and co-teacher regularization strategies. 
Remarkably, Model E trained solely with the augmentation strategy outperforms FACT, even without the Tail Interaction. 
The outcome further validates the effectiveness of our proposed Two-step High-pass Filter strategy.

In addition, as demonstrated in Table \ref{tab:ab30}, the exclusion of any component from the Tail Interaction module leads to a decline in generalization performance, whereas the combination of all four components yields the optimal results.
These findings indicate the indispensability of each individual part for achieving effective generalization, with dual normalization emerging as a favorable selection in the module.
Furthermore, the specific form and extent of normalization present promising avenues for future investigations.

\begin{table*}[!t]
\footnotesize
\centering
\caption{Ablation studies ($\%$) on PACS dataset.
The best performance is marked as \textbf{bold}.
\textbf{HP}, \textbf{PS}, \textbf{AS}, and \textbf{TI} indicate \textbf{H}igh-\textbf{P}ass filter augmentation, \textbf{P}hase \textbf{S}caling, \textbf{A}mplitude \textbf{S}caling, and \textbf{T}ail \textbf{I}nteraction, respectively.
A, C, P, and S indicate the target domain of Art painting, Cartoon, Photo, and Sketch.
The backbone is ResNet-18
}
\begin{tabular}{l|ccc|c|cccc>{\columncolor{gray!15}}c}
\specialrule{0em}{1pt}{1pt}\hline\toprule[0.5pt]\specialrule{0em}{1pt}{1pt}
\textbf{Methods}  & {\hspace{0.35em}\textbf{HP}} & {\hspace{0.3em}\textbf{PS}} & {\hspace{0.25em}\textbf{AS}} & {\hspace{0.25em}\textbf{TI}} & {\textbf{A}} & {\textbf{C}} & {\textbf{P}} & {\textbf{S}} & \cellcolor{gray!15}{\textbf{Avg.}} \\ \specialrule{0em}{1pt}{1pt}\hline\specialrule{0em}{1pt}{1pt}
Baseline & \textbf{ -}                      & \textbf{ -}                      & \textbf{ -}                      & \textbf{ -}                       & $77.63\pm0.00$            & $76.77\pm0.00$           & $95.85\pm0.00$           & $69.50\pm0.00$            & 79.94                   \\ \specialrule{0em}{1pt}{1pt}\hline\specialrule{0em}{1pt}{1pt}
Model A  & \hspace{0.35em}\checkmark                      & \textbf{ -}                      & \textbf{ -}                      & \textbf{ -}                       &  $78.28\pm0.83$                     & $78.76\pm1.35$                      & $96.00\pm0.24$                      &  $79.04\pm0.99$                      & 83.02                         \\
Model B  & \hspace{0.35em}\checkmark                      & \textbf{ -}                      & \textbf{ -}                      & \hspace{0.35em}\checkmark                       &  $78.52\pm0.74$                     &  $79.11\pm1.41$                     &  $\textbf{97.45}\pm0.18$                     & $82.12\pm1.06$                      & 84.30                         \\
Model C  & \hspace{0.35em}\checkmark                      & \textbf{ -}                      & \hspace{0.35em}\checkmark                      & \hspace{0.35em}\checkmark                       & $80.46\pm0.60$                      &   $79.03\pm0.67$                   &   $97.31\pm0.38$                    &  $82.28\pm1.03$                      &84.77                        \\
Model D  & \hspace{0.35em}\checkmark                      & \hspace{0.35em}\checkmark                      & \textbf{ -}                      & \hspace{0.35em}\checkmark                       & $79.92\pm0.36$                      &  $78.11\pm0.72$                    &  $96.74\pm0.39$                    & $82.59\pm1.26$                       &84.34                          \\
Model E  & \hspace{0.35em}\checkmark  & \hspace{0.35em}\checkmark  & \hspace{0.35em}\checkmark  & \textbf{ -} & $81.37\pm0.51$                 & $79.16\pm0.80$                 & $96.28\pm0.22$                 &$83.49\pm0.83$                  & 85.08                                          \\ \specialrule{0em}{1pt}{1pt}\hline\specialrule{0em}{1pt}{1pt}
+ All modules  & \hspace{0.35em}\checkmark                      & \hspace{0.35em}\checkmark                      & \hspace{0.35em}\checkmark                      & \hspace{0.35em}\checkmark                                 & $\textbf{83.51}\pm0.65$          & $\textbf{81.34}\pm1.09$          & $97.20\pm0.41$    & $\textbf{85.11}\pm0.90$         & \textbf{86.79}                   
                \\ 
\specialrule{0em}{1pt}{1pt}\hline\toprule[0.5pt]\specialrule{0em}{1pt}{1pt}
\end{tabular}
\label{tab:ab0}
\vspace{-5pt} 
\end{table*}

\begin{table}[htbp]
\centering
\small
\caption{Ablation studies ($\%$) for the Tail Interaction module on PACS dataset.
A, C, P, and S indicate the target domain of Art painting, Cartoon, Photo, and Sketch.
The best performance is marked as \textbf{bold}.
The frequency restriction module is built on Two-step High-pass Filter.
The backbone is ResNet-18
}
\resizebox{\linewidth}{!}{
\begin{tabular}{l|cccc>{\columncolor{gray!15}}c}
\hline
\toprule[0.5pt]
\specialrule{0em}{1pt}{1pt}
\textbf{Methods}                        & \textbf{A} & \textbf{C} & \textbf{P} & \textbf{S} & \textbf{Avg.}   \\ \specialrule{0em}{1pt}{1pt}\hline \specialrule{0em}{1pt}{1pt}
Baseline                      & 77.63\tiny$\pm0.00$                      & 76.77\tiny$\pm0.00$      & 95.85\tiny$\pm0.00$      & 69.50\tiny$\pm0.00$      & 79.94             \\
\specialrule{0em}{1pt}{1pt}\hline \specialrule{0em}{1pt}{1pt}
w/o $I_k$         & 81.42\tiny$\pm0.30$          & 79.22\tiny$\pm0.71$          & 96.53\tiny$\pm0.08$    & 83.89\tiny$\pm0.85$         & 85.27                   \\ 
w/o $I_v$      & 81.55\tiny$\pm0.44$          & 79.96\tiny$\pm0.64$          & 96.89\tiny$\pm0.17$    & 83.83\tiny$\pm1.11$         & 85.56                   \\ 
w/o $N_{s}$        & 82.09\tiny$\pm0.38$          & 80.41\tiny$\pm0.82$          & 96.77\tiny$\pm0.12$    & 84.35\tiny$\pm0.76$         & 85.91                   \\ 
w/o $N_{l_1}$    & 83.03\tiny$\pm0.71$          & 80.86\tiny$\pm1.13$          & 96.85\tiny$\pm0.10$    & 84.70\tiny$\pm0.94$         & 86.36        
           \\
\specialrule{0em}{1pt}{1pt}\hline \specialrule{0em}{1pt}{1pt}
Ours & \textbf{83.51}\tiny$\pm0.65$          & \textbf{81.34}\tiny$\pm1.09$          & \textbf{97.20}\tiny$\pm0.41$    & \textbf{85.11}\tiny$\pm0.90$         & \textbf{86.79}  \\
           \specialrule{0em}{1pt}{1pt}\hline
\toprule[0.5pt]
\end{tabular}}
\vspace{-0.5cm}
\label{tab:ab30}
\end{table}

\subsection{Further Analysis}
\label{sec:analysis}
\begin{figure}[htbp]
\setlength{\belowcaptionskip}{-2cm}
 \setlength{\abovecaptionskip}{-0.1cm}
  \centering
\includegraphics[width=0.5\textwidth]{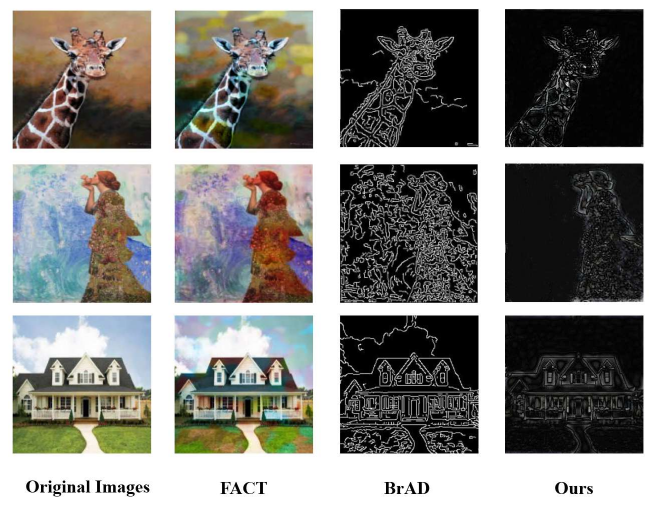}
   \caption{Visualization comparison between FACT (Amplitude Mix), BrAD, and our augmentation (Two-step High-pass Filter).}
   \label{fig:41}
\end{figure}

\noindent{{\bf Comparison of results between our proposed method, FACT and BrAD.}} 
We compare the performance of using our proposed method, FACT and BrAD.  
The Table \ref{tab:brad} demonstrates that, with the sole utilization of data augmentation, the performance improvement resulting from our proposed augmentation strategy surpasses that of FACT and BrAD.
Further improvements are achieved by incorporating the Tail Interaction module. 
This demonstrates the superiority of our proposed augmentation strategy and the effectiveness of the Tail Interaction module.
In addition, we provide a visual comparison of our proposed augmentation with FACT and BrAD in \reffig{fig:41}.  
As we mentioned in \refsec{sec:hpf}, our method ensures the integrity and correctness of generalizable features in generated images. In contrast, the linear interpolation of amplitude spectrum in FACT leads to image distortion and introduce unrealistic features, while the reliance on edge models in BrAD overlooks other important semantic information or introduce irrelevant background edges, reducing the expressiveness of generated images.

\begin{table}[htbp]
\centering
\normalsize
\caption{Comparison ($\%$) between augmentation methods proposed by FACT, BrAD, and us on different datasets (in $\%$). 
The best performance is marked as \textbf{bold} for different backbones, respectively.
TI, AS, AM, GK, and THF indicate Tail Interaction, Amplitude Swap, Amplitude Mix, Gaussian Kernel, and Two-step High-pass Filter, respectively
}   
\resizebox{\linewidth}{!}{
\begin{tabular}{l|cccc}
\hline
\toprule[0.5pt]
\specialrule{0em}{1pt}{1pt}
\textbf{Methods}                        & \textbf{PACS} & \hspace{20pt}\textbf{VLCS} & \textbf{\hspace{6pt}Office-Home} & \textbf{Digits-DG}  \\ \specialrule{0em}{1pt}{1pt}\hline \specialrule{0em}{1pt}{1pt}
\multicolumn{1}{l|}{}     & \multicolumn{3}{c}{\textit{ResNet-18}} & \textit{\begin{tabular}[c]{@{}c@{}}Backbone in\\ \cite{14,45}\end{tabular}} \\
\specialrule{0em}{1pt}{1pt} \hline 
\specialrule{0em}{1pt}{1pt}
Baseline                      & 79.94                     & \hspace{20pt}72.48      & \hspace{6pt}64.72      & 73.70                 \\
+ Canny (BrAD)        & 83.75  &\hspace{20pt}74.90      &\hspace{6pt}65.58    &  78.21     
           \\
+ Canny (BrAD)\&TI         & 84.02       & \hspace{20pt}75.25     &\hspace{6pt}65.90      & 80.74       
           \\
+ HED (BrAD)                     & 84.52      & \hspace{20pt}74.98  & \hspace{6pt}65.73  & 80.00     
\\
+ HED (BrAD)\&TI         & 86.11      &\hspace{20pt}76.52   & \hspace{6pt}67.01  &  82.43    
           \\
+ AS (FACT)            & 82.80             & \hspace{20pt}74.15     & \hspace{6pt}65.03    & 77.86    
\\
+ AS (FACT)\&TI     & 83.18      & \hspace{20pt}76.30     &\hspace{6pt}66.42     & 80.57                 
\\
+ AM (FACT)                      &  83.42            &\hspace{20pt}74.88     &\hspace{6pt}65.42     & 78.00    
           \\
+ AM (FACT)\&TI        & 84.13  &\hspace{20pt}76.59     & \hspace{6pt}66.70    &  81.05  
           \\
+ GK                      &84.83   &  \hspace{20pt}75.24    & \hspace{6pt}66.38     &81.75       
\\
+ GK\&TI      & 86.77     &\hspace{20pt}77.79     & \hspace{6pt}67.40    & 83.40     
           \\ 
+ THF     &85.08             &\hspace{20pt}76.04      & \hspace{6pt}66.40    &81.81       
\\
+ THF\&TI     &  \textbf{86.79}   &\hspace{20pt}\textbf{78.11}      & \hspace{6pt}\textbf{67.59}   & \textbf{83.70}   
           \\
\specialrule{0em}{1pt}{1pt}\hline \specialrule{0em}{1pt}{1pt}
\multicolumn{1}{l|}{}     & \multicolumn{4}{c}{\textit{GFNet-H-Ti}}  \\ \specialrule{0em}{1pt}{1pt} \hline
\specialrule{0em}{1pt}{1pt}
Baseline                      & 87.75         &\hspace{20pt}78.60      & \hspace{6pt}70.38     &   87.30              \\
+ Canny (BrAD)        &88.38   &\hspace{20pt}79.20      &\hspace{6pt}72.03    & 88.96      
           \\
+ Canny (BrAD)\&TI         & 89.45       &\hspace{20pt}80.12     &\hspace{6pt}72.98      &91.37        
           \\
+ HED (BrAD)                     &88.64       &\hspace{20pt}79.17   &\hspace{6pt}72.33   & 89.06     
\\
+ HED (BrAD)\&TI         & 89.47      &\hspace{20pt}80.13  & \hspace{6pt}73.27  &  91.50   
           \\
+ AS (FACT)            & 88.06             & \hspace{20pt}79.05     &\hspace{6pt}71.84     &  88.38   
\\
+ AS (FACT)\&TI     &89.31       & \hspace{20pt}79.69     & \hspace{6pt}72.56    & 90.15                  
\\
+ AM (FACT)                      & 88.12             &\hspace{20pt}79.20      & \hspace{6pt}71.99    & 88.93    
           \\
+ AM (FACT)\&TI        & 89.26            &\hspace{20pt}80.02      & \hspace{6pt}72.73    & 90.28   
           \\
+ GK                      &88.84   & \hspace{20pt}79.23    &  \hspace{6pt}72.48    & 89.52       
\\
+ GK\&TI                      & 89.49             &\hspace{20pt}80.13      &\hspace{6pt}73.29     & 91.60     
           \\ 
+ THF                     &  89.29            & \hspace{20pt}79.44     & \hspace{6pt}72.67    & 90.39      
\\
+ THF\&TI                      & \textbf{89.55}             & \hspace{20pt}\textbf{80.21}     & \hspace{6pt}\textbf{74.08}    &   \textbf{92.30} 
           \\
           \specialrule{0em}{1pt}{1pt}\hline
\toprule[0.5pt]
\end{tabular}}
\label{tab:brad}
\end{table}

\noindent{\bf Comparison of results between self-attention and Tail Interaction.} 
We evaluate the influence of employing self-attention, which exhibits limitations in domain generalization.
As illustrated in Table \ref{tab:self}, the combination of self-attention with our proposed frequency restriction strategies yields inferior results compared to utilizing the Tail Interaction module we introduce. 
These results provide quantitative evidence of the constraints of the self-attention mechanism in model generalization.
This discrepancy arises from the fact that the Tail Interaction module learns domain-invariant representations across all samples from various domains, whereas self-attention solely captures interactions within individual samples, disregarding the interactions between different samples.

\begin{table}[t]
\centering
\normalsize
\caption{Comparison ($\%$) between self-attention and Tail Interaction on PACS.
The best performance is marked as \textbf{bold} for different backbones, respectively.
A, C, P, and S indicate the target domain of Art painting, Cartoon, Photo, and Sketch.
THF, GK, SA, and TI indicate Two-step High-pass Filter, Gaussian Kernel, Self-Attention, and Tail Interaction, respectively
}
\resizebox{\linewidth}{!}{
\begin{tabular}{l|cccc>{\columncolor{gray!15}}c}
\hline
\toprule[0.5pt]
\specialrule{0em}{1pt}{1pt}
\textbf{Methods}                        & \textbf{A} & \textbf{C} & \textbf{P} & \textbf{S} & \textbf{Avg.}   \\ \specialrule{0em}{1pt}{1pt}\hline \specialrule{0em}{1pt}{1pt}
\multicolumn{1}{l|}{}     & \multicolumn{5}{c}{\textit{ResNet-18}}  \\ \specialrule{0em}{1pt}{1pt} \hline
\specialrule{0em}{1pt}{1pt}
Baseline                      & 77.63\tiny$\pm0.00$                      & 76.77\tiny$\pm0.00$      & 95.85\tiny$\pm0.00$      & 69.50\tiny$\pm0.00$      & 79.94             \\
+ GK\&SA                      & 80.32\tiny$\pm0.76$                      &78.53\tiny$\pm0.81$      & 96.92\tiny$\pm0.08$      & 83.20\tiny$\pm0.64$      & 84.74  
           \\
+ THF\&SA                      & 81.34\tiny$\pm0.65$                      & 79.19\tiny$\pm0.68$      & 96.22\tiny$\pm0.34$      & 83.37\tiny$\pm0.91$      &  85.03 
           \\
+ GK\&TI                      & \textbf{83.73}\tiny$\pm0.48$                      & \textbf{81.40}\tiny$\pm0.92$      & 96.89\tiny$\pm0.33$      & 85.07\tiny$\pm0.71$      & 86.77  
           \\ 
+ THF\&TI                      & 83.51\tiny$\pm0.65$                      & 81.34\tiny$\pm1.09$      & \textbf{97.20}\tiny$\pm0.41$      & \textbf{85.11}\tiny$\pm0.90$      & \textbf{86.79}      
           \\
\specialrule{0em}{1pt}{1pt}\hline \specialrule{0em}{1pt}{1pt}
\multicolumn{1}{l|}{}     & \multicolumn{5}{c}{\textit{GFNet-H-Ti}}  \\ \specialrule{0em}{1pt}{1pt} \hline
\specialrule{0em}{1pt}{1pt}
 Baseline                & 91.20\tiny$\pm0.85$        & 82.28\tiny$\pm1.45$     & \textbf{99.03}\tiny$\pm0.14$      & 78.48\tiny$\pm1.28$   &87.75          \\
+ GK\&SA                        & 90.59\tiny$\pm0.77$        & 82.89\tiny$\pm0.85$     & 98.68\tiny$\pm0.12$      & 83.12\tiny$\pm0.64$   &88.82
           \\
+ THF\&SA                          & \textbf{91.58}\tiny$\pm0.61$        & 82.72\tiny$\pm1.12$     & 98.81\tiny$\pm0.18$      & 84.06\tiny$\pm1.10$   & 89.29
           \\
+ GK\&TI                           & 91.05\tiny$\pm0.14$        & \textbf{83.23}\tiny$\pm1.02$     & 98.67\tiny$\pm0.01$      & 85.01\tiny$\pm0.47$   &89.49 
           \\
+ THF\&TI                           & 91.07\tiny$\pm0.15$        & 83.17\tiny$\pm1.71$     & 98.76\tiny$\pm0.05$      & \textbf{85.19}\tiny$\pm1.06$   &\textbf{89.55}
           \\
           \specialrule{0em}{1pt}{1pt}\hline
\toprule[0.5pt]
\end{tabular}}
\label{tab:self}
\end{table}

\begin{table}[htbp]
\centering
\normalsize
\caption{Comparison ($\%$) between different number of $I_k$ and $I_v$ on different datasets.
The augmentation strategy is Two-step High-pass Filter.
The best performance is marked as \textbf{bold} for different backbones, respectively
}
\resizebox{\linewidth}{!}{
\begin{tabular}{l|cccc}
\hline
\toprule[0.5pt]
\specialrule{0em}{1pt}{1pt}
\textbf{Methods}                        & \textbf{PACS} & \textbf{\hspace{20pt}VLCS} & \textbf{\hspace{6pt}Office-Home} & \textbf{Digits-DG}  \\ \specialrule{0em}{1pt}{1pt}\hline \specialrule{0em}{1pt}{1pt}
\multicolumn{1}{l|}{}     & \multicolumn{3}{c}{\textit{ResNet-18}} & \textit{\begin{tabular}[c]{@{}c@{}}Backbone in\\ \cite{14,45}\end{tabular}} \\
\specialrule{0em}{1pt}{1pt} \hline 
\specialrule{0em}{1pt}{1pt}
Baseline                      & 79.94                     & \hspace{20pt}72.48      & \hspace{6pt}64.72      & 73.70                 \\
one $I_k$ and $I_v$        & \textbf{86.79}  &\hspace{20pt}\textbf{78.11}      &\hspace{6pt}67.59   &  83.70     
           \\
two $I_k$ and $I_v$         & 86.71       & \hspace{20pt}78.02     &\hspace{6pt}67.63      & \textbf{83.84}       
           \\
three $I_k$ and $I_v$                     & 86.65      & \hspace{20pt}77.89  & \hspace{6pt}\textbf{67.70}  & 83.81     
\\
\specialrule{0em}{1pt}{1pt}\hline \specialrule{0em}{1pt}{1pt}
\multicolumn{1}{l|}{}     & \multicolumn{4}{c}{\textit{GFNet-H-S}}  \\ \specialrule{0em}{1pt}{1pt} \hline
\specialrule{0em}{1pt}{1pt}
Baseline                      & 87.75         &\hspace{20pt}78.60      & \hspace{6pt}70.38     &   87.30              \\
one $I_k$ and $I_v$        &\textbf{89.55}  &\hspace{20pt}\textbf{80.21}      &\hspace{6pt}\textbf{74.08}    &  \textbf{92.30}     
           \\
two $I_k$ and $I_v$         & 89.50       & \hspace{20pt}80.13     &\hspace{6pt}73.96      & 92.17       
           \\
three $I_k$ and $I_v$                     & 88.98      & \hspace{20pt}80.07  & \hspace{6pt}73.94  & 91.05     
\\
           \specialrule{0em}{1pt}{1pt}\hline
\toprule[0.5pt]
\end{tabular}}
\label{tab:multii}
\end{table}

\noindent{\bf The effects of using multiple $I_v$ and $I_k$ in Tail Interaction module.}
In Table \ref{tab:multii}, we conduct experiments with different quantities of $I_k$ and $I_v$ within the Tail Interaction module. 
In the experiments, we select ResNet-18 with the smallest network size and GFNet-H-S with the largest network size as backbones for training among four network architectures (ResNet-18 (12M), ResNet-50 (26M), GFNet-H-Ti (15M), and GFNet-H-S (32M)).
From the table, it can be observed that as the number of $I_k$ and $I_v$ gradually increases, the performance exhibits different rules for different datasets and network architectures. 
\textit{When using ResNet-18 as the backbone, employing multiple $I_k$ and $I_v$ improves performance on the Office-Home dataset but degrades performance on PACS and VLCS. On the Digits-DG dataset, performance initially improves and then declines. 
When GFNet-H-S is used as the backbone, performance decreases on all these datasets.}
The $I_k$ and $I_v$ are two units in the Tail Interaction module, implemented using two linear layers.
Using multiple $I_k$ and $I_v$ is equivalent to increasing the channel dimension, which may enhance the model's expressive ability but also increase the risk of overfitting. The specific effect depends on the characteristics and difficulty of the dataset, as well as the characteristics of the network.
For the case of ResNet-18 as the backbone, using multiple $I_k$ and $I_v$ can improve the performance on the Office-Home (67.59 $\rightarrow$ 67.63 (+0.04) $\rightarrow$ 67.70 (+0.07)) dataset. This is because the Office-Home dataset has small differences between categories, containing 65 fine-grained categories, which makes it challenging for the model to distinguish between different categories, requiring more channel dimensions to enhance the expressive ability and differentiate these categories. However, on the PACS (86.79 $\rightarrow$ 86.71 (-0.08) $\rightarrow$ 86.65 (-0.06)) and VLCS (78.11 $\rightarrow$ 78.02 (-0.09) $\rightarrow$ 77.89 (-0.13)) datasets, using multiple $I_k$ and $I_v$ may decrease performance due to large domain differences, requiring more domain-invariant features to adapt to target domains. Using multiple $I_k$ and $I_v$ lead to overfitting to the source domain features and weaken the learning of domain-invariant features. The sample size of Digits-DG dataset is small, and the image clarity and contrast are low. On the Digits-DG (83.70 $\rightarrow$ 83.84 (+0.14) $\rightarrow$ 83.81 (-0.03)) dataset, performance initially improves and then decreases. We believe that using multiple $I_k$ and $I_v$ enhances the model's robustness to some extent, but excessive $I_k$ and $I_v$ may lead to overfitting to noise and details, losing the overall features of the digits.
For GFNet-H-S as the backbone, the performance on these datasets decreases. GFNet-H-S itself is a model that uses self-attention (global filter), having strong expressive and generalization capabilities. Increasing the number of $I_k$ and $I_v$ results in excessive model parameters and computational complexity, potentially disrupting the balance and optimization of the original self-attention, thereby reducing model performance. 
Choosing an appropriate number of $I_k$ and $I_v$ involves considering several factors, including:
\begin{itemize}
    \item \textbf{Characteristics of the dataset:} Different datasets have distinct feature distributions, complexities, and diversities. For datasets with large domain shift, a smaller number of $I_k$ and $I_v$ suffices to learn domain-invariant features. Conversely, datasets with complex and fine-grained categories require more $I_k$ and $I_v$ to enhance the model's expressive and generalization abilities.
    \item \textbf{Difficulty of the task:} For image classification tasks, focusing on global features requires fewer $I_k$ and $I_v$ for good performance. In contrast, tasks like object detection or semantic segmentation, which require attention to local features, benefit from more $I_k$ and $I_v$ to enhance network expressiveness and robustness.
    \item \textbf{Structure of the model:} If other parts of the model are already complex or have a large number of parameters, reducing the number of $I_k$ and $I_v$ is necessary to lower computational and memory costs and avoid overfitting. Simpler models with fewer parameters benefit from an increased number of $I_k$ and $I_v$ to improve performance.
    \item \textbf{Methodology of the training:} The use of regularization or data augmentation techniques allows for the use of more $I_k$ and $I_v$ to enhance network expressiveness and generalization. Conversely, pruning or quantization methods necessitates fewer $I_k$ and $I_v$ to reduce computational and memory costs and improve the efficiency.
\end{itemize}

\noindent{\bf The universality of the proposed modules.}
We assess the effectiveness of our proposed modules on several advanced DG methods and other MLPs on PACS. 
It is worth noting that the DG modules increase the number of parameters by about 60,000 (only the Tail Interaction module adds parameters). 
Specifically, we insert the proposed frequency restriction and the Tail Interaction module into EFDMix, SWAD, Mixer-B, gMLP-S, and ResMLP-S. 
Please note that when our proposed method is combined with other augmented methods, we feed the original images, results from other augmented methods, and our augmentation results into the network for learning. We keep the bachsize of the original images the same in our experiments. Moreover, EFDMix is a feature-level data augmentation technique as it generates new samples in the feature space, thereby enhancing data diversity. In contrast, our data augmentation method operates at the image-level, and the two methods are not mutually exclusive. 
As demonstrated in Table \ref{tab:addPACS}, all models achieve further improvements in average performance after embedding our modules. 
Furthermore, the baseline result provided by GFNet-H-Ti is the strongest among the listed MLPs, with the fewest parameters and the best final average performance of the model after embedding our two modules.

\begin{table*}[htbp]
\footnotesize
\centering
\caption{DG accuracy ($\%$) on PACS.
A, C, P, and S indicate the target domain of Art painting, Cartoon, Photo, and Sketch.
G in brackets means using Gaussian Kernel, and T means using Two-step High-pass Filter.
The text in blue indicates the number of parameters.
SWAD{*} is reproduced by \cite{75} based on the released code of \cite{91}
}
\setlength{\tabcolsep}{4mm}{
\begin{tabular}{l|c|cccc>{\columncolor{gray!15}}c}
\hline
\toprule[0.5pt]
\specialrule{0em}{1pt}{1pt}
\textbf{Methods}               & \textbf{Venue}                & \textbf{A} & \textbf{C} & \textbf{P} & \textbf{S} & \textbf{Avg.}  \\ \specialrule{0em}{1pt}{1pt}\hline \specialrule{0em}{1pt}{1pt}
\multicolumn{1}{c|}{\textit{}} & \multicolumn{1}{l|}{}         & \multicolumn{5}{c}{\textit{ResNet-18}}                                                 \\ \specialrule{0em}{1pt}{1pt} \hline
\specialrule{0em}{1pt}{1pt}
DeepAll \cite{5} \textcolor{blue}{(12M)}                       & CVPR'21                     & 77.63\tiny$\pm$0.00      & 76.77\tiny$\pm$0.00      & 95.85\tiny$\pm$0.00      & 69.50\tiny$\pm$0.00      & 79.94         \\
+ Our modules (G)                    & This paper      & 83.73\tiny$\pm$0.48      
& 81.40\tiny$\pm$0.92      
& 96.89\tiny$\pm$0.33      
& 85.07\tiny$\pm$0.71      
& 86.77 (\textcolor{red}{+6.83})         \\
+ Our modules (T)                    & This paper      & 83.51\tiny$\pm$0.65      
& 81.34\tiny$\pm$1.09      
& 97.20\tiny$\pm$0.41      
& 85.11\tiny$\pm$0.90      
& 86.79 (\textcolor{red}{+6.85}) \\ \specialrule{0em}{1pt}{1pt}\hline\specialrule{0em}{1pt}{1pt}
EFDMix \cite{74} \textcolor{blue}{(12M)}                       & CVPR'22                       & 83.90\tiny$\pm$0.40      & 79.40\tiny$\pm$0.70      & 96.80\tiny$\pm$0.40      & 75.00\tiny$\pm$0.70      & 83.90              \\
+ Our modules (G)                    & This paper  & 85.20\tiny$\pm$0.30      & 79.60\tiny$\pm$0.70      & 96.60\tiny$\pm$0.40      & 80.00\tiny$\pm$0.40      & 85.40 (\textcolor{red}{+1.50}) \\
+ Our modules (T)                    & This paper  & 85.30\tiny$\pm$0.40      & 79.60\tiny$\pm$0.60      & 96.70\tiny$\pm$0.70      & 80.20\tiny$\pm$0.40      & 85.50 (\textcolor{red}{+1.60}) \\ \specialrule{0em}{1pt}{1pt}\hline\specialrule{0em}{1pt}{1pt}
{SWAD{*}} \cite{91} \textcolor{blue}{(12M)}                    & CVPR'22                       & 83.10\tiny$\pm$1.50      & 75.90\tiny$\pm$0.90      & 95.60\tiny$\pm$0.60      & 77.10\tiny$\pm$2.40      & 82.90         \\   
+ Our modules (G)                    & This paper  & 83.04\tiny$\pm$1.30      & 77.83\tiny$\pm$1.00      & 96.48\tiny$\pm$0.30      & 76.69\tiny$\pm$1.40      & 83.51 (\textcolor{red}{+0.61}) \\
+ Our modules (T)                    & This paper & 81.94\tiny$\pm$1.10      & 77.08\tiny$\pm$1.00      & 96.23\tiny$\pm$0.40      & 78.79\tiny$\pm$1.70      & 83.86 (\textcolor{red}{+0.96}) \\
 \specialrule{0em}{1pt}{1pt}\hline\specialrule{0em}{1pt}{1pt} & \multicolumn{1}{l|}{}         & \multicolumn{5}{c}{\textit{MLPs}}                   \\ \specialrule{0em}{1pt}{1pt}\hline\specialrule{0em}{1pt}{1pt}

Mixer-B \cite{28} \textcolor{blue}{(59M)}                     & This paper                       & 85.11\tiny$\pm$0.70      & 77.78\tiny$\pm$1.12      & 94.60\tiny$\pm$0.26      & 65.53\tiny$\pm$1.19      & 80.76               \\
+ Our modules (G)                    & This paper 
& 85.40\tiny$\pm$0.62      
& 78.37\tiny$\pm$0.94      
& 95.22\tiny$\pm$0.35      
& 79.17\tiny$\pm$1.23      
& 84.54 (\textcolor{red}{+3.78}) \\
+ Our modules (T)                    & This paper
& 85.29\tiny$\pm$0.83      
& 79.09\tiny$\pm$0.66      
& 95.54\tiny$\pm$0.28      
& 79.93\tiny$\pm$1.54      
& 84.96 (\textcolor{red}{+4.20}) \\ \specialrule{0em}{1pt}{1pt}\hline\specialrule{0em}{1pt}{1pt}
gMLP-S \cite{92} \textcolor{blue}{(20M)}                    & This paper                       & 86.59\tiny$\pm$0.75     & 81.02\tiny$\pm$1.53      & 97.64\tiny$\pm$0.18      & 72.12\tiny$\pm$1.16    &84.34      \\   
+ Our modules (G)                    & This paper
& 84.48\tiny$\pm$1.09      
& 82.06\tiny$\pm$0.74      
& 98.38\tiny$\pm$0.03      
& 80.76\tiny$\pm$1.31      
& 86.42 (\textcolor{red}{+2.08})\\
+ Our modules (T)                    & This paper 
& 85.00\tiny$\pm$0.55      
& 81.86\tiny$\pm$0.61      
& 97.78\tiny$\pm$0.14      
& 80.84\tiny$\pm$0.97      
& 86.37 (\textcolor{red}{+2.03})\\
\specialrule{0em}{1pt}{1pt}\hline\specialrule{0em}{1pt}{1pt}
ResMLP-S \cite{93} \textcolor{blue}{(15M)}                    & This paper                       & 85.47\tiny$\pm$0.71     & 78.60\tiny$\pm$1.46      & 97.19\tiny$\pm$0.27      & 72.53\tiny$\pm$2.03    &83.45      \\   
+ Our modules (G)                    & This paper 
& 85.50\tiny$\pm$0.64      
& 79.79\tiny$\pm$0.50      
& 98.33\tiny$\pm$0.32      
& 80.62\tiny$\pm$0.88      
& 86.06 (\textcolor{red}{+2.61})\\
+ Our modules (T)                    & This paper
& 85.36\tiny$\pm$1.08      
& 79.51\tiny$\pm$0.67      
& 97.96\tiny$\pm$0.15      
& 80.77\tiny$\pm$1.23      
& 85.90 (\textcolor{red}{+2.45})\\
\specialrule{0em}{1pt}{1pt}\hline\specialrule{0em}{1pt}{1pt}
GFNet-H-Ti \cite{30} \textcolor{blue}{(15M)}                   & \multicolumn{1}{c|}{This paper} & 91.20\tiny$\pm$0.85      & 82.28\tiny$\pm$1.45      & 99.03\tiny$\pm$0.14      & 78.48\tiny$\pm$1.28      & 87.75              \\
+ Our modules (G)                           & This paper                    & 91.05\tiny$\pm$0.14      & 83.23\tiny$\pm$1.02      & 98.67\tiny$\pm$0.01      & 85.01\tiny$\pm$0.47      & 89.49 (\textcolor{red}{+1.74})                 \\
+ Our modules (T)                           & This paper                    & 91.07\tiny$\pm$0.15      & 83.17\tiny$\pm$1.71      & 98.76\tiny$\pm$0.05      & 85.19\tiny$\pm$1.06      & 89.55 (\textcolor{red}{+1.80})         \\ \specialrule{0em}{1pt}{1pt}\hline
\toprule[0.5pt]
\end{tabular}}
\vspace{-15pt}
\label{tab:addPACS}
\end{table*}

\begin{figure}[htbp]
\vspace{-12pt}
\setlength{\abovecaptionskip}{-0.005cm}
  \centering  \includegraphics[width=1\linewidth]{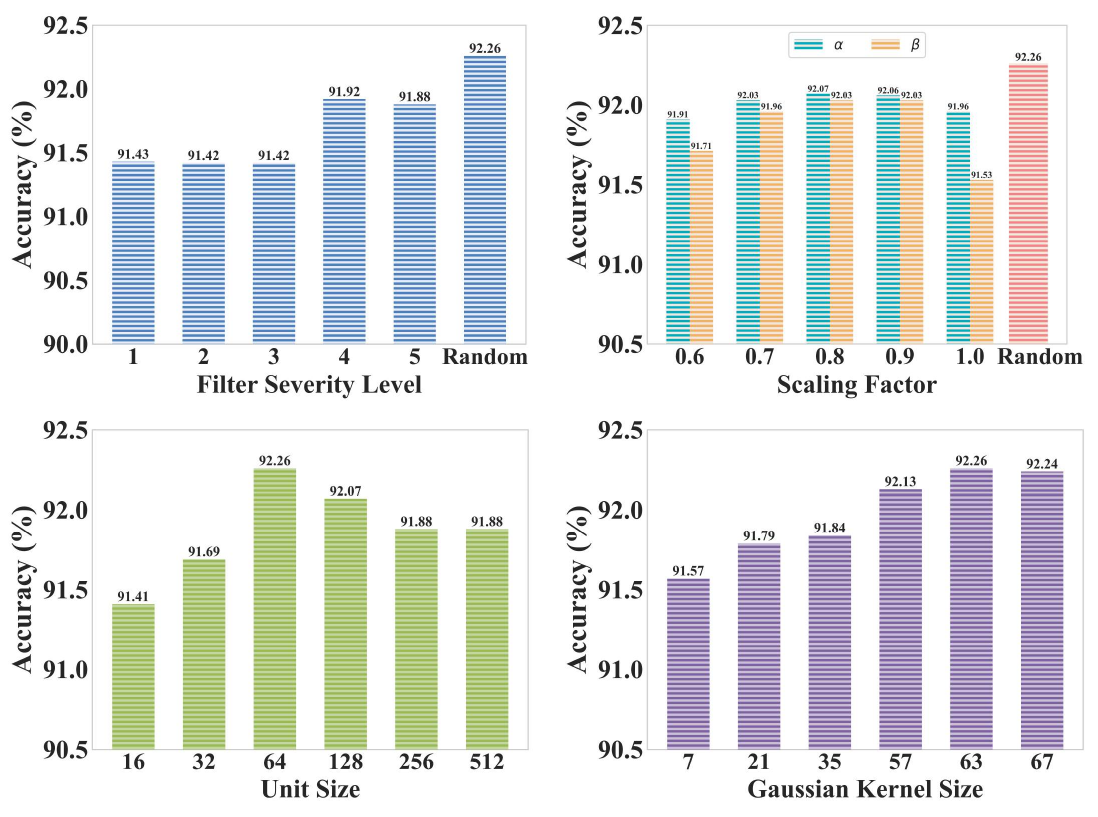}
   \caption{Effects of hyper-parameters including filter severity level, scaling factor, interaction unit size, and Gaussian Kernel size.
   The experiments are conducted on Digits-DG with GFNet-H-Ti as the backbone.}
   \label{fig:hist}
\vspace{-5pt}
\end{figure}

\noindent{\bf The effects of hyper-parameters.}
\reffig{fig:hist} illustrates the impact of hyper-parameters on model performance. 
(1) Using random filter severity levels in the high-pass filter results in higher accuracy than using one of the five fixed values.
We believe adjusting the filter severity level adaptively for samples from different domains can be more effective than using random severity levels.
(2) The Tail Interaction unit with a small size is effective, and the complexity is linear in the number of pixels.
(3) Using a small Gaussian Kernel can work well and does not impact model training speed.

\begin{figure}[htbp]
%%\vspace{-1pt}
\setlength{\abovecaptionskip}{-0.05cm}
\setlength{\belowcaptionskip}{-2.2cm}
  \centering  \includegraphics[width=1\linewidth]{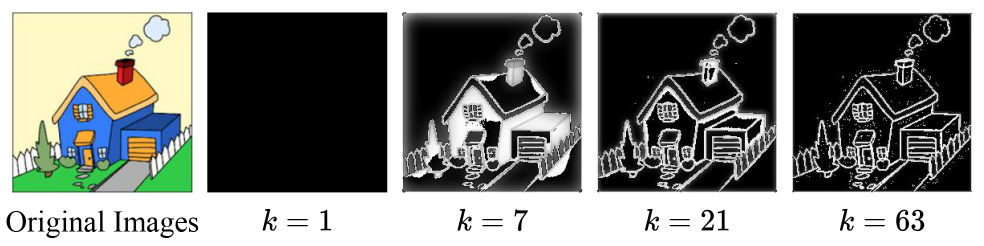}
   \caption{Visualization examples of filtering via Gaussian Kernel with different kernel sizes, retaining semantics.}
   \label{fig:gauu}
   \vspace{-8pt}
\end{figure}

\noindent{\bf Image visualization after filtering via Gaussian Kernel.}
We show the difference between using different Gaussian Kernel sizes in \reffig{fig:gauu}.
The resulting high frequency image contains domain-invariant semantic concepts, removing the background and other visual features like color and illumination that is unrelated to the identity and structure.

\begin{figure}[htbp]
\vspace{-5pt}
\setlength{\abovecaptionskip}{-0.1cm}
\setlength{\belowcaptionskip}{-2cm}
  \centering
\includegraphics[width=1\linewidth]{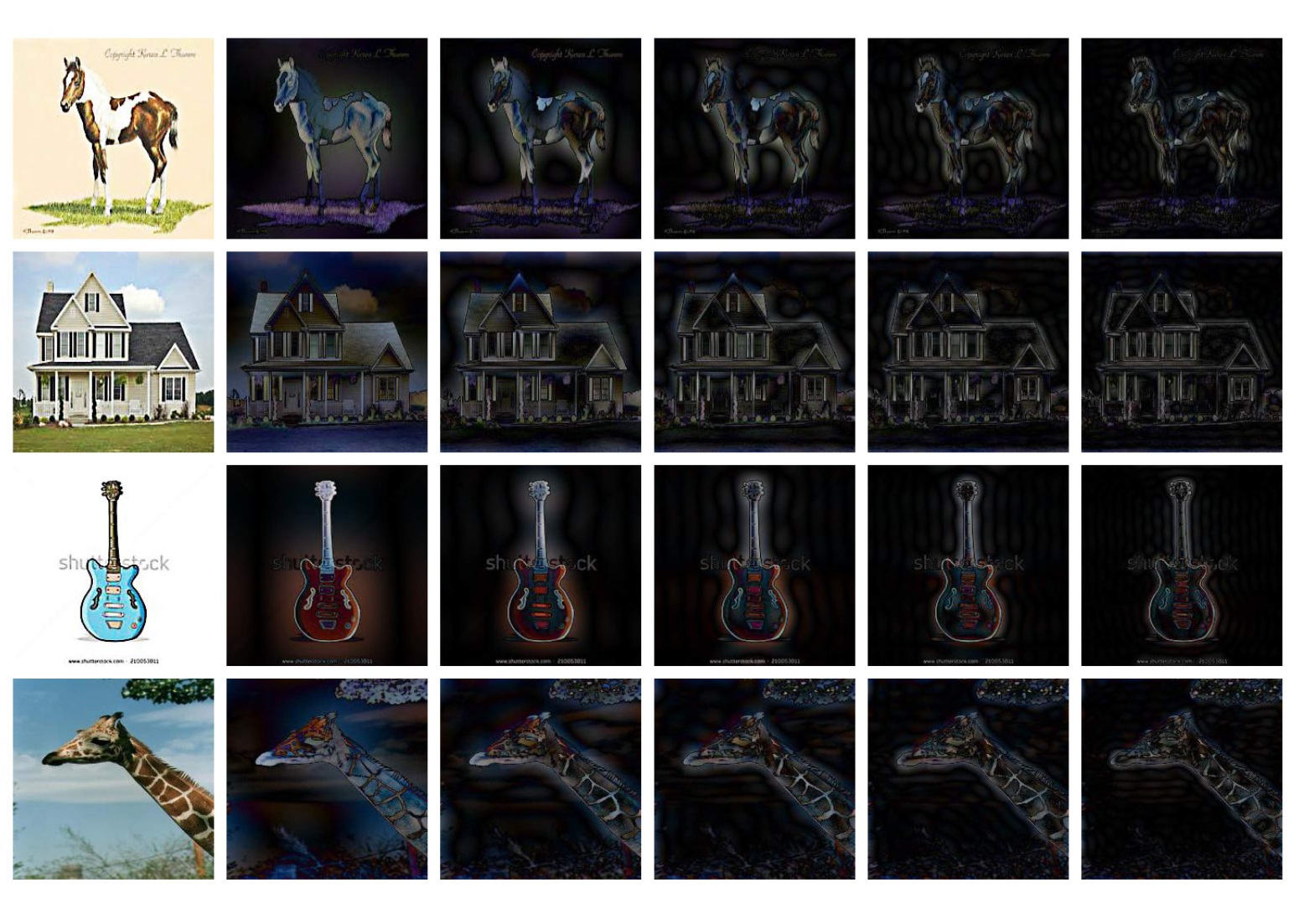}
   \caption{Visualization examples of high-pass filtered for five severity levels. 
   The level increases from left to right, and the value of the diameter $d$ is 0, 2.24, 4.48, 6.72, 8.96, and 11.2, respectively.}
   \label{fig:hp0}
   \vspace{-7pt}
\end{figure}

\noindent{\bf Image visualization after Two-step High-pass Filter.} 
In Eq. (\ref{eq:5}), $d$ is proportional to the dimensions of the input images.
Since we use 224$\times$224 images, the respective values of d are in the set 224$\times$\{0.01, 0.02, 0.03, 0.04, 0.05\}, where
$224\times0.05=11.2$ is the maximum filtering level.
We show the difference between the severity levels in \reffig{fig:hp0}.
These images alleviate the lack of images that do not contain spurious correlations in the dataset. 
The expectation is that the model acquires complete semantics as robust, domain-invariant representations from these images. 
%free of significant disturbance from other features.
% Other visual features unrelated to labels cannot significantly disturb it.

\begin{figure}[htbp]
\vspace{-2.5mm}
\setlength{\belowcaptionskip}{-1cm}
\centering
\subfloat[Images before scaling]{
\begin{minipage}[t]{0.23\linewidth}
\includegraphics[width=0.75in]{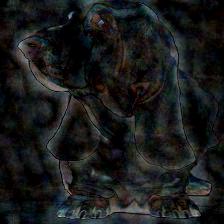}
%\caption{fig1}
\end{minipage}%
\begin{minipage}[t]{0.23\linewidth}
\includegraphics[width=0.75in]{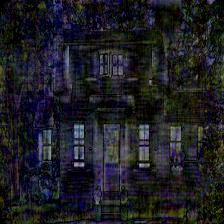}
%\caption{fig2}
\end{minipage}%
\label{fig:s1}
}%
\subfloat[Images after scaling]{
\begin{minipage}[t]{0.22\linewidth}
\includegraphics[width=0.75in]{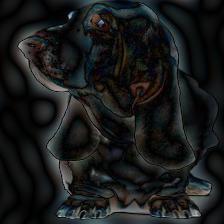}
%\caption{fig2}
\end{minipage}
\begin{minipage}[t]{0.21\linewidth}
\includegraphics[width=0.75in]{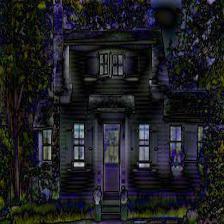}
%\caption{fig2}
\end{minipage}
\label{fig:s2}
}%
\centering
\caption{Visualization examples before and after scaling amplitude and phase.
After scaling, the lines of the objects become apparent, and they no longer look ``noisy".}
\vspace{-10pt}
\end{figure}

\noindent{\bf The effects of scaling amplitude and phase on noise reduction.}
As mentioned before, the high frequency component inevitably retains the noise during high-pass filtering.
In \reffig{fig:s1}, we visually depict the noise retained in the high-pass filtered images. 
Additionally, in \reffig{fig:s2}, we show that scalings of amplitude and phase aid in noise reduction.
The results from these visualizations, combined with the ablation studies results presented in Table \ref{tab:ab0}, prove that noise interferes with the model learning semantics as domain-invariant representations to a certain extent, thereby reducing the generalization performance of the model.
These visualized samples further validate the effectiveness and rationality of this module design.

\begin{figure*}[htbp]
\vspace{-6pt}
\setlength{\abovecaptionskip}{-0.1cm}
  \centering
\includegraphics[width=1\linewidth]{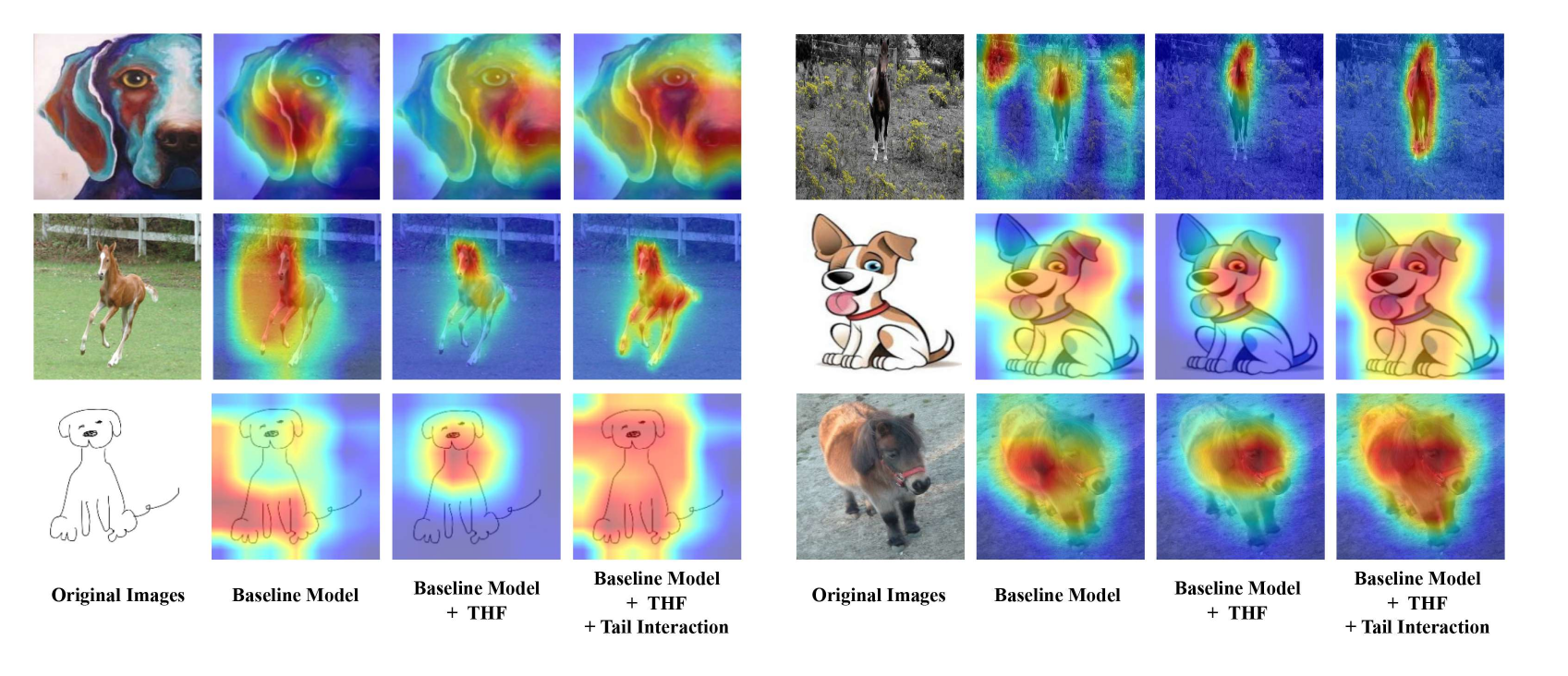}
  \caption{GradCAM visualizations of the baseline model and our models. The experiments are based on ResNet-18. THF indicates Two-step High-pass Filter.}
   \label{fig:gradcam}
\end{figure*}

\noindent{\bf Visual explanations (GradCAM) for a comprehensive understanding of the two proposed modules.} 
In \reffig{fig:gradcam}, the visual explanations highlight the impact of our modules on the baseline model. The Two-step High-pass Filter refines the model's focus, reducing interference from spurious correlations (such as the grass background in the second row, second column), while the Tail Interaction module enhances the model's capacity to grasp intricate and comprehensive features. This dual-module integration helps the model effectively explore and enhance the utilization of the common features among samples from diverse domains. The model can better focus on domain-invariant representation of objects, ultimately boosting its generalization performance.

\begin{figure}[htbp]
  \centering
  \setlength{\abovecaptionskip}{-0.1cm}
  \setlength{\belowcaptionskip}{-4cm}
\includegraphics[width=1\linewidth]{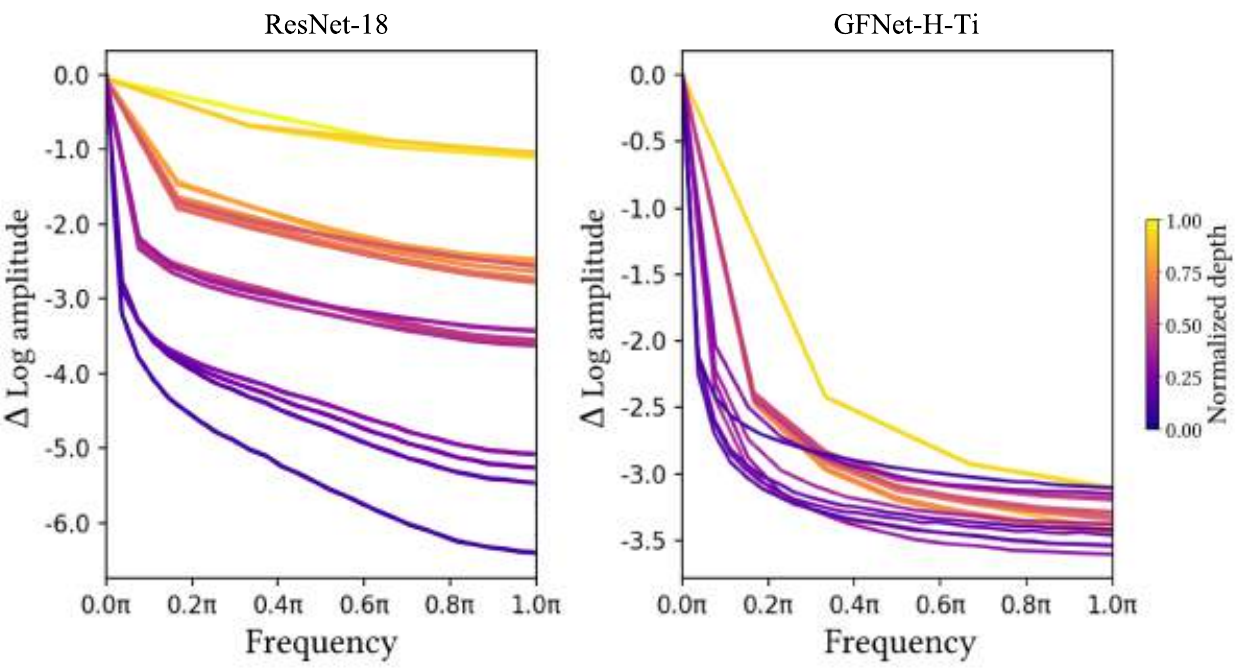}
   \caption{Relative log amplitudes of Fourier transformed feature maps.
   $\Delta$ Log amplitude of high frequency components is the difference between the log amplitude at normalized frequency $0.0\pi$ (center) and at $1.0\pi$ (boundary).}
   \label{fig:fre}
\end{figure}

\noindent{\bf Intuition and insight into modules.}
Each module we design embodies its own intuition and insights, and they are not trivially embedded into the models.
The Fourier analysis of the feature maps \cite{90} in \reffig{fig:fre} shows that both CNNs and MLPs decrease high frequency components when inputting different frequencies to them. 
CNNs exhibit a weaker response to high frequency components in lower layers, which improves in higher layers. 
MLPs attenuate high frequency components in all layers, but the reduction is less pronounced than in CNNs. 
Both models attenuate high frequency components more strongly than low frequency components. 
The significance of high frequency components for semantic information makes them critical for model generalization. 
These findings indicate the necessity to compensate for the attenuation of high frequency information and to encourage the models to learn more domain-invariant information from high frequency components.
Moreover, in the outlook section of \cite{70}, Ishaan Gulrajani and David Lopez-Paz wrote, ``We think of data augmentation as \textit{feature removal}. If the practitioner is lucky and performs the data augmentations that cancel the spurious correlations varying from domain to domain, then out-of-distribution performance should improve."
Fortunately, our frequency restriction module, helps the model get rid of spurious correlations and acquire domain-invariant representations, resulting in new state-of-the-art. 
Tail Interaction considers potential correlations among all samples from all domains, making up for the fact that the global filter layer in GFNet only learns interaction among spatial locations within a single sample and ignores the interaction between different samples.
These two modules have proven to be both simple and effective, bringing performance improvement to the model.
% For example, only employing our augmentation module based on Two-step High-pass Filter outperforms the GFNet-H-Ti baseline by 1.54$\%$ on PACS.

\section{Conclusion}
\label{sec:conclusion}
In this paper, we explore the generalization performance of the model toward learning domain-invariant features by disentangling spurious correlations and enhancing potential correlations. 
From the sample perspective, we design a frequency restriction module to extract semantics information.
From the feature perspective, we introduce the Tail Interaction to consider potential correlations among all samples across multi-domains. 
Whether embedded in CNNs or MLPs, the two modules can help improve the performance.
% Our models based on CNNs and MLPs outperform the state-of-the-arts. 
Visualizations highlight the effectiveness of our method and the reasonableness of some settings, validating the importance of relevant and broad correlations for domain-invariant representations.

% \section*{Acknowledgment}
% This work was supported by the NSFC Program (62222604, 62206052, 62192783), China Postdoctoral Science Foundation (2021M690609), Jiangsu Natural Science Foundation (BK20210224), CCF-Lenovo Bule Ocean Research Fund.

{\small
\balance
\bibliographystyle{ieee_fullname}
\bibliography{egbib}
}

\newpage
\setcounter{section}{0}
\setcounter{subsection}{0}
\section{Content}
\label{sec:intro}

The content of this supplementary material involves:
(1) Visualization examples of filtered and scaled images: \refsec{sec:vis}.
(2) Additional experiments, including the introduction of DomainBed benchmark: \refsec{sec:dom}, implementation details: \refsec{sec:imp}, experimental results: \refsec{sec:res}, ablation studies: \refsec{sec:abl}, and further analysis: \refsec{sec:fur}.

\section{More Visualization Examples}
\label{sec:vis}
\begin{figure}[htbp]
\setlength{\belowcaptionskip}{-0.3cm}
  \centering
   \includegraphics[width=1\linewidth]{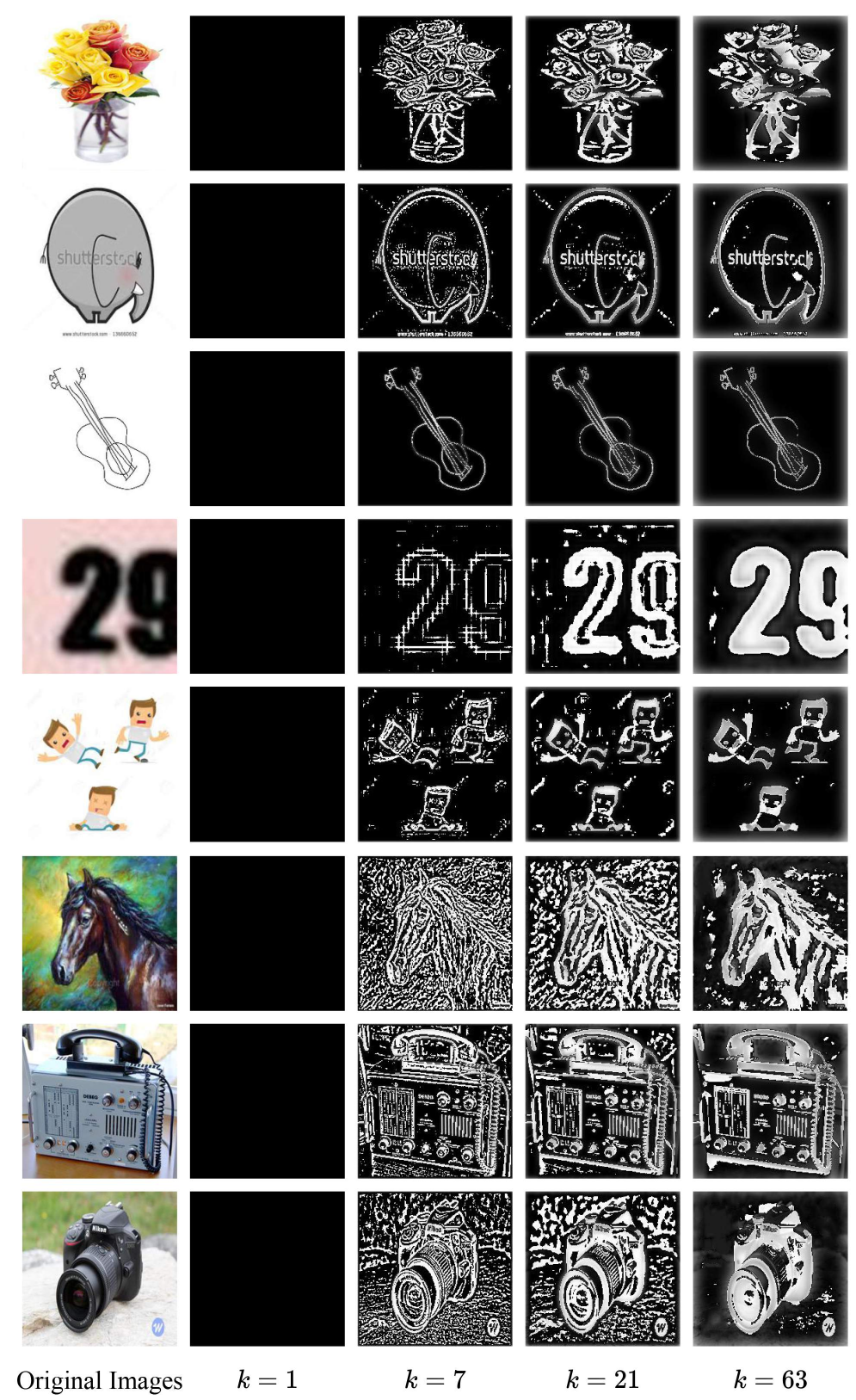}
   \caption{More visualization examples of the grayscale high frequency components as domain-invariant representations.
   We employ Gaussian Kernel with different kernel sizes $k$.}
   \label{fig:gau}
\end{figure}

\textbf{Gaussian Kernel filtered images.}
Intuitively, the low frequency components of an image capture information such as color and illumination, whereas the high frequency components correspond to sharp edges and basic shapes of objects. 
To extract the high frequency components of the original images, we applied Gaussian blur with various kernel sizes, and the resulting images are shown in \reffig{fig:gau}. 

\begin{figure}[htbp]
  \centering
   \includegraphics[width=1\linewidth]{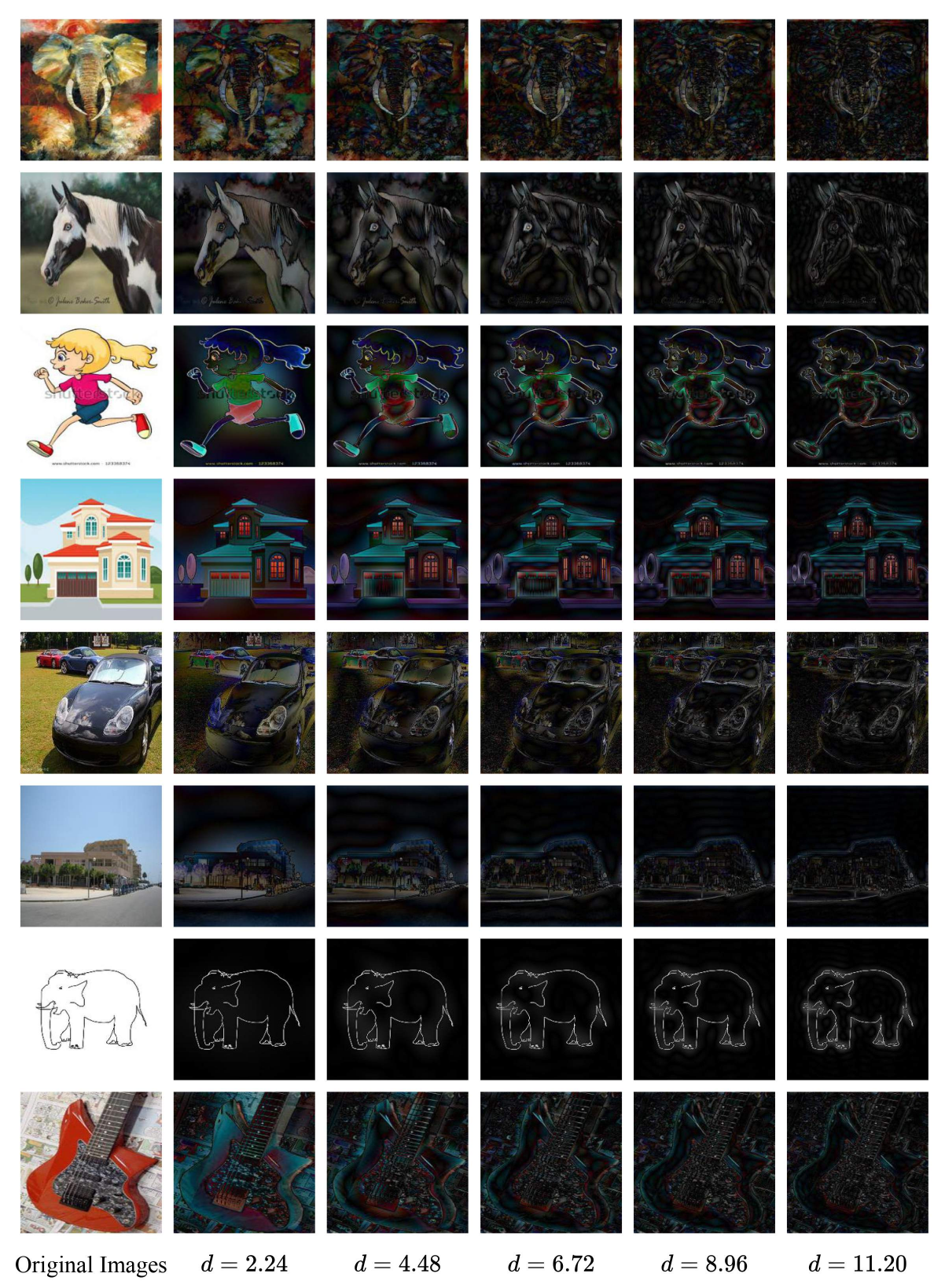}
   \caption{More visualization examples of high-pass filtered with five severity levels. 
   The severity level increases from left to right, filtering out more low frequency components.}
   \label{fig:hp}
\end{figure}

\textbf{High-pass filtered images.}
We present more visualization examples of high-pass filtered images with varying levels of severity, along with their corresponding original images, in \reffig{fig:hp}.
The visual structures of different objects are generally preserved in the high frequency components, allowing for the retention of complete semantic information.

\vspace{-5pt}

\begin{figure}[h]
  \centering
   \includegraphics[width=1\linewidth]{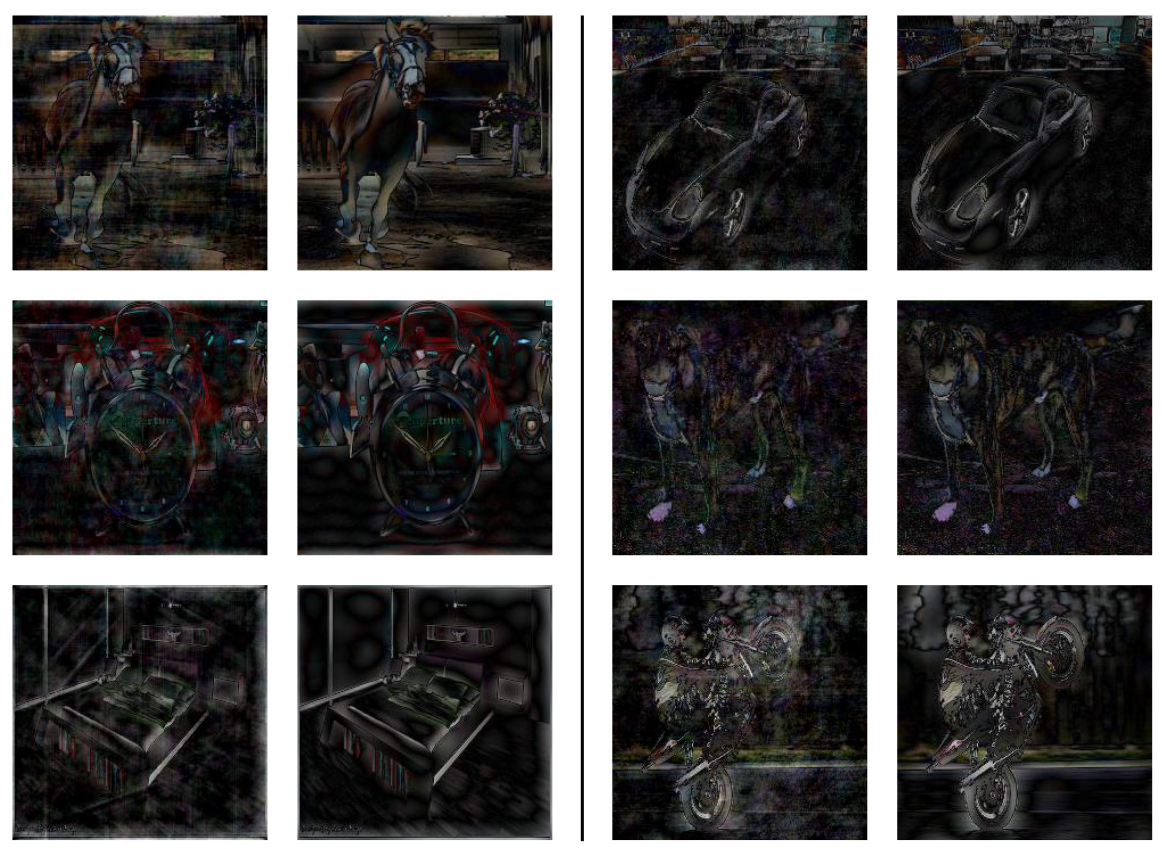}
   \caption{More visualization examples before and after scaling the amplitude and phase components. 
   Each pair of images consists of the original image on the left and the scaled image on the right, demonstrating a reduction in noise compared to the original version.}
   \label{fig:sca}
\end{figure}

\textbf{Images before and after scaling.}
The effect of scaling the amplitude and phase components on image noise reduction is demonstrated in \reffig{fig:sca}. 
Each pair of images shows the image before scaling on the left and the scaled image on the right. 
The results show that these scalings significantly impact reducing noise in the images.
For instance, in the first row, the outline of the car appears blurry in the third column image, while the lines and edges of the car are clearer in the fourth column image of the same row.

\begin{table*}[t]
\footnotesize
\centering
\caption{Out-of-domain accuracy ($\%$) with train-validation selection criterion.
The best performance is marked as \textbf{bold} for different backbones, respectively.
TerraInc indicates TerraIncognita.
G in brackets means using Gaussian Kernel, and T means using Two-step High-pass Filter.
The backbones of MLPs are GFNet-H-Ti and GFNet-H-S, abbreviated as Ti and S
}
\setlength{\tabcolsep}{4mm}{
\begin{tabular}{l|c|ccccc>{\columncolor{gray!15}}c}
\specialrule{0em}{1pt}{1pt}\hline\toprule[0.5pt]\specialrule{0em}{1pt}{1pt}
\textbf{Methods}               & \textbf{Venue}        & \textbf{PACS} & \textbf{VLCS} & \textbf{Office-Home} & \textbf{TerraInc}& \textbf{DomainNet} & \textbf{Avg.} \\ \specialrule{0em}{1pt}{1pt}\hline\specialrule{0em}{1pt}{1pt}
\multicolumn{1}{c|}{\textit{}} & \multicolumn{1}{l|}{} & \multicolumn{5}{c}{\textit{ResNet-50}}                                  \\ \specialrule{0em}{1pt}{1pt}\hline\specialrule{0em}{1pt}{1pt}
IRM \cite{77} & Arxiv'19 & 83.5\tiny$\pm$0.8           & 78.5\tiny$\pm$0.5             & 64.3\tiny$\pm$2.2          & 47.6\tiny$\pm$0.8 & 33.9\tiny$\pm$2.8         & 61.6         \\
ERM  \cite{121}    & ICLR'20                   & 85.5\tiny$\pm$0.2            &77.5\tiny$\pm$0.4            & 66.5\tiny$\pm$0.3             & 46.1\tiny$\pm$1.8         & 40.9\tiny$\pm$0.1         & 63.3          \\
SWAD \cite{91}                    & NeurIPS'21               & 88.1\tiny$\pm$0.1           & 79.1\tiny$\pm$0.1             & 70.6\tiny$\pm$0.2          & 50.0\tiny$\pm$0.3 &\textbf{46.5}\tiny$\pm$0.1         & 66.9         \\
SelfReg \cite{122}                    & ICCV'21               & 85.6\tiny$\pm$0.4           & 77.8\tiny$\pm$0.9             & 67.9\tiny$\pm$0.7          & 47.0\tiny$\pm$0.3 &41.5\tiny$\pm$0.2         & 64.0         \\
Fish \cite{123}              & ICLR'22               & 85.5\tiny$\pm$0.3           & 77.8\tiny$\pm$0.3             & 68.6\tiny$\pm$0.4         & 45.1\tiny$\pm$1.3         & 42.7\tiny$\pm$0.2   &63.9          \\
MIRO \cite{124}                & ECCV'22               & 85.4\tiny$\pm$0.4           & 79.0\tiny$\pm$0.0            & 70.5\tiny$\pm$0.4          & 50.4\tiny$\pm$1.1        & 44.3\tiny$\pm$0.2  &65.9          \\
MVRML \cite{55}                   & ECCV'22               & \textbf{88.3}\tiny$\pm$0.3           & 77.9\tiny$\pm$0.2            & 71.3\tiny$\pm$0.1          & 51.0\tiny$\pm$0.6         & 45.7\tiny$\pm$0.1 &66.8  \\
GVRT \cite{125}                   & ECCV'22               & 85.1\tiny$\pm$0.3           & 79.0\tiny$\pm$0.2            & 70.1\tiny$\pm$0.1          & 48.0\tiny$\pm$0.2         & 44.1\tiny$\pm$0.1 &65.2   \\
PGrad \cite{126}                   & ICLR'23               & 85.1\tiny$\pm$0.3           & 79.3\tiny$\pm$0.3            & 69.3\tiny$\pm$0.1          & 49.0\tiny$\pm$0.3         & 41.0\tiny$\pm$0.1 &64.7   \\
Balance \cite{127}                   & ICLR'23               & 86.1\tiny$\pm$0.4           & 76.1\tiny$\pm$0.3            & 67.1\tiny$\pm$0.4          & 48.0\tiny$\pm$1.7         & 42.6\tiny$\pm$1.0 &64.0   \\
% SAGM \cite{61}                  & CVPR'23               & 86.6\tiny$\pm$0.2           & 80.0\tiny$\pm$0.3             & 70.1\tiny$\pm$0.2          & 48.8\tiny$\pm$0.9 &45.0\tiny$\pm$0.2         & 66.1         \\
DAC-P \cite{128}                  & CVPR'23               & 85.6\tiny$\pm$0.5           & 77.0\tiny$\pm$0.6             & 69.5\tiny$\pm$0.1          & 49.8\tiny$\pm$0.2 &43.8\tiny$\pm$0.3         & 65.1         \\
Ours (G) & This paper  
&86.8\tiny$\pm$0.3  &79.3\tiny$\pm$0.6 &70.9\tiny$\pm$0.3 &50.7\tiny$\pm$0.9   &46.2\tiny$\pm$0.2  &66.8 \\
Ours (T) & This paper  
&87.3\tiny$\pm$0.2 
&\textbf{79.7}\tiny$\pm$0.5 &\textbf{71.4}\tiny$\pm$0.0 &\textbf{51.1}\tiny$\pm$1.2
&\textbf{46.5}\tiny$\pm$0.1 
&\textbf{67.2}
\\ \specialrule{0em}{1pt}{1pt}\hline\specialrule{0em}{1pt}{1pt}
\multicolumn{1}{c|}{\textit{}} &                       & \multicolumn{5}{c}{\textit{MLPs}}                                                \\ \specialrule{0em}{1pt}{1pt}\hline\specialrule{0em}{1pt}{1pt}
% SDViT \cite{45}                   & Arxiv'22              & 86.3\tiny$\pm0.2$         & 78.9\tiny$\pm0.4$            & 71.5\tiny$\pm0.2$         & 44.3\tiny$\pm1.0$        & 45.8\tiny$\pm0.0$ &65.3           \\ %\specialrule{0em}{1pt}{1pt}\hline\specialrule{0em}{1pt}{1pt}
% \multicolumn{1}{c|}{\textit{}} &                       & \multicolumn{5}{c}{\textit{MLPs}}                                                \\ \specialrule{0em}{1pt}{1pt}\hline\specialrule{0em}{1pt}{1pt}
Baseline (Ti)                   &  This paper            & 87.5\tiny$\pm$0.1           & 79.0\tiny$\pm$0.0    &71.5\tiny$\pm$0.2          &  47.8\tiny$\pm$0.6       & 46.0\tiny$\pm$0.2    & 66.4    \\ 
Ours (G)                        & This paper            &  88.6\tiny$\pm$0.4          & 79.2\tiny$\pm$0.6            & 72.0\tiny$\pm$0.1          & 51.1\tiny$\pm$0.6         &  46.3\tiny$\pm$0.0   &67.4         \\
Ours (T)                        & This paper            &  88.5\tiny$\pm$0.2          & 79.9\tiny$\pm$0.2            & 73.1\tiny$\pm$0.4          &51.4\tiny$\pm$0.7         &  \textbf{47.8}\tiny$\pm$0.2  &68.1         \\
Baseline (S)                   &  This paper            &  88.2\tiny$\pm$0.2         &   79.5\tiny$\pm$1.0   &  71.9\tiny$\pm$0.1        & 48.9\tiny$\pm$0.4  &46.2\tiny$\pm$0.0        & 67.0         \\
Ours (G)  & This paper  
&\textbf{88.7}\tiny$\pm$0.4 
&79.7\tiny$\pm$0.3 
&73.1\tiny$\pm$0.2
&51.4\tiny$\pm$1.0
&47.5\tiny$\pm$0.2  
&68.1  \\
Ours (T)  & This paper  
&\textbf{88.7}\tiny$\pm$0.3  
&\textbf{80.6}\tiny$\pm$0.5 
&\textbf{73.8}\tiny$\pm$0.4  
&\textbf{51.6}\tiny$\pm$1.1 
&\textbf{47.8}\tiny$\pm$0.4   
&\textbf{68.5}  \\
\specialrule{0em}{1pt}{1pt}\hline\toprule[0.5pt]\specialrule{0em}{1pt}{1pt}
\end{tabular}
}
\label{tab:dbed}
\end{table*}

\begin{table*}[!t]
\vspace{-5pt}
\footnotesize
\centering
\caption{Ablation studies ($\%$) on PACS dataset.
The best performance is marked as \textbf{bold}.
\textbf{HP}, \textbf{PS}, \textbf{AS}, and \textbf{TI} indicate \textbf{H}igh-\textbf{P}ass filter augmentation, \textbf{P}hase \textbf{S}caling, \textbf{A}mplitude \textbf{S}caling, and \textbf{T}ail \textbf{I}nteraction, respectively.
A, C, P, and S indicate the target domain of Art painting, Cartoon, Photo, and Sketch.
The backbone is GFNet-H-Ti
}
\begin{tabular}{l|ccc|c|cccc>{\columncolor{gray!15}}c}
\specialrule{0em}{1pt}{1pt}\hline\toprule[0.5pt]\specialrule{0em}{1pt}{1pt}
\textbf{Methods}  & {\hspace{0.35em}\textbf{HP}} & {\hspace{0.3em}\textbf{PS}} & {\hspace{0.25em}\textbf{AS}} & {\hspace{0.25em}\textbf{TI}} & {\textbf{A}} & {\textbf{C}} & {\textbf{P}} & {\textbf{S}} & \cellcolor{gray!15}{\textbf{Avg.}} \\ \specialrule{0em}{1pt}{1pt}\hline\specialrule{0em}{1pt}{1pt}
Baseline & \textbf{ -}                      & \textbf{ -}                      & \textbf{ -}                      & \textbf{ -}                       & $91.20\pm0.85$            & $82.28\pm1.45$           & $\textbf{99.03}\pm0.14$           & $78.48\pm1.28$            & 87.75                    \\ \specialrule{0em}{1pt}{1pt}\hline\specialrule{0em}{1pt}{1pt}
Model A  & \hspace{0.35em}\checkmark                      & \textbf{ -}                      & \textbf{ -}                      & \textbf{ -}                       &  $90.75\pm0.41$                     & $81.81\pm2.56$                      & $98.85\pm0.08$                      &  $84.16\pm1.16$                      & 88.89                         \\
Model B  & \hspace{0.35em}\checkmark                      & \textbf{ -}                      & \textbf{ -}                      & \hspace{0.35em}\checkmark                       &  $91.01\pm0.58$                     &  $82.97\pm1.54$                     &  $98.94\pm0.16$                     & $83.25\pm2.18 $                      &  89.04                        \\
Model C  & \hspace{0.35em}\checkmark                      & \textbf{ -}                      & \hspace{0.35em}\checkmark                      & \hspace{0.35em}\checkmark                       & $91.64\pm0.73$                      &   $81.79\pm1.42$                   &   $98.92\pm0.24$                    &  $84.42\pm0.90$                      &   89.19                       \\

Model D  & \hspace{0.35em}\checkmark                      & \hspace{0.35em}\checkmark                      & \textbf{ -}                      & \hspace{0.35em}\checkmark                       & $90.92\pm1.02$                      &  $82.28\pm1.69$                    &  $98.86\pm0.18$                    & $84.14\pm1.27$                       & 89.05                         \\
Model E  & \hspace{0.35em}\checkmark  & \hspace{0.35em}\checkmark  & \hspace{0.35em}\checkmark  & \textbf{ -} & $\footnotesize\textbf{91.66}\pm0.59$                 & $82.70\pm1.67$                 & $98.79\pm0.17$                 & $84.00\pm1.25$                  & 89.29                                           \\ \specialrule{0em}{1pt}{1pt}\hline\specialrule{0em}{1pt}{1pt}
+ All modules   & \hspace{0.35em}\checkmark                      & \hspace{0.35em}\checkmark                      & \hspace{0.35em}\checkmark                      & \hspace{0.35em}\checkmark                                 & $91.07\pm0.85$          & $\footnotesize\textbf{83.17}\pm1.71$          & $98.76\pm0.18$    & $\footnotesize\textbf{85.19}\pm1.06$         & \textbf{89.55}                    \\ 
\specialrule{0em}{1pt}{1pt}\hline\toprule[0.5pt]\specialrule{0em}{1pt}{1pt}
\end{tabular}
\label{tab:ab} 
\end{table*}

\begin{table}[t]
\vspace{-10pt}
\centering
\normalsize
\caption{Ablation studies ($\%$) for the Tail Interaction module on PACS dataset.
A, C, P, and S indicate the target domain of Art painting, Cartoon, Photo, and Sketch.
The best performance is marked as \textbf{bold}.
The frequency restriction module is built on Two-step High-pass Filter.
The backbone is GFNet-H-Ti
}
\resizebox{\linewidth}{!}{
\begin{tabular}{l|cccc>{\columncolor{gray!15}}c}
\hline
\toprule[0.5pt]
\specialrule{0em}{1pt}{1pt}
\textbf{Methods}                        & \textbf{A} & \textbf{C} & \textbf{P} & \textbf{S} & \textbf{Avg.}   \\ \specialrule{0em}{1pt}{1pt}\hline \specialrule{0em}{1pt}{1pt}
Baseline                      & 91.20\tiny$\pm0.85$                      & 82.28\tiny$\pm1.45$      & \textbf{99.03}\tiny$\pm0.14$      & 78.48\tiny$\pm1.28$      & 87.75             \\
Ours & 91.07\tiny$\pm0.85$          & \textbf{83.17}\tiny$\pm1.71$          & 98.76\tiny$\pm0.18$    & \textbf{85.19}\tiny$\pm1.06$         & \textbf{89.55}  \\
w/o $I_k$  & \textbf{91.52}\tiny$\pm0.67$          & 83.04\tiny$\pm1.01$          & 98.80\tiny$\pm0.09$    & 84.11\tiny$\pm1.15$         & 89.37                   \\ 
w/o $I_v$  & 90.88\tiny$\pm0.56$          & 83.06\tiny$\pm1.24$          & 98.79\tiny$\pm0.12$    & 84.87\tiny$\pm1.00$         & 89.40                    \\ 
w/o $N_{s}$    & 91.28\tiny$\pm0.44$          & 83.06\tiny$\pm1.16$          & 98.91\tiny$\pm0.12$    & 84.55\tiny$\pm1.24$         & 89.45                   \\ 
w/o $N_{l_1}$                                & 91.46\tiny$\pm0.57$          & 83.08\tiny$\pm1.33$          & 98.92\tiny$\pm0.23$    & 84.50\tiny$\pm0.94$         &89.49                    \\ 
           \specialrule{0em}{1pt}{1pt}\hline
\toprule[0.5pt]
\end{tabular}}
\vspace{-10pt}
\label{tab:ab3}
\end{table}

\vspace{-12pt}
\section{Additional Experiments}
\label{sec:exp}
\subsection{DomainBed Benchmark}
\label{sec:dom}
Following \cite{70}, we further employ the DomainBed evaluation protocols for an extensive comparison on various benchmarks: PACS (4 domains, 7 classes, and 9,991 images), VLCS (4 domains, 5 classes, and 10,729 images), Office-Home (4 domains, 65 classes, and 15,588 images), TerraIncognita \cite{72} (4 domains, 10 classes, and 24,788 images), and DomainNet \cite{73} (6 domains, 345 classes, and 586,575 images).
DomainBed serves as a testbed for domain generalization, implementing consistent experimental protocols across state-of-the-art methods to ensure a fair and reliable comparison.

\subsection{Implementation Details}
\label{sec:imp}
For experiments on DomainBed, we optimize the models using AdamW with a learning rate of 5e-4 and no weight decay when the backbone is GFNet. 
ImageNet pre-trained ResNet-50 is also employed as the initial weight and optimized by Adam \cite{120} optimizer with a learning rate of 5e-5.
We keep data augmentation strategy and other hyper-parameters, such as batch size, dropout rate, and training steps, consistent with the default configuration in DomainBed.
We choose a domain as the target domain and use the remaining domains as the training domain where 20$\%$ samples are used for validation and model selection.
Each out-of-domain performance we report averages three different runs with different train-validation splits.
Other details are the same as in the main paper.

\subsection{Results}
\label{sec:res}
We provide exhaustive out-of-domain performance comparisons on five DG benchmarks in Table \ref{tab:dbed}.
The proposed model (T) based on GFNet-H-S significantly improves performance on every benchmark dataset, achieving 68.5$\%$ average performance.
Compared with the state-of-the-art methods based on ResNet-50, our model (T) achieves the best performances in all benchmarks except PACS. 
Especially, our model (T) based on ResNet-50 remarkably outperforms previous methods: +3.1$\%$ on TerraIncognita (GVRT: 48.0$\%$$\rightarrow$51.1$\%$) and +2.8$\%$ on Office-Home (Fish: 68.6$\%$$\rightarrow$71.4$\%$). 
TerraIncognita focuses on the generalization to new environments, including the backgrounds and overall lighting conditions, precisely the type of spurious correlations we expect to eliminate.
Considering the extensive experiment setup with five datasets and 22 target domains, the results demonstrate the effectiveness of our modules to the diverse visual data types.

\subsection{Ablation Studies}
\label{sec:abl}
We conduct supplementary ablation studies to investigate the role of each module in our framework based on MLPs, as shown in Table \ref{tab:ab}. 
The backbone is GFNet-H-Ti.
Model A is trained only using the data augmentation based on the High-Pass (HP) filter from the baseline. 
Based on model A, we add the Tail Interaction (TI) module to obtain model B, which improves over model A. 
Our complete network, which incorporates Phase Scaling (PS) and Amplitude Scaling (AS), outperforms all other variants. 
The results suggest the crucial role of scaling the amplitude and phase in reducing noise interference and improving model performance.
We also validate the necessity of the scaling of phase or amplitude by using only HP, TI, and the other, resulting in models C and D, respectively.
We can observe that neither model C nor D outperforms the complete network, suggesting the effectiveness of incorporating both scalings to reduce noise during acquiring domain-invariant representations.
We further create model E by excluding Tail Interaction, and the performance drops, showing that this module plays an indispensable role.

Furthermore, as shown in Table \ref{tab:ab3}, the removal of any part from the Tail Interaction module results in a deterioration of generalization performance, while the integration of all four elements achieves the best result.
These results highlight the essential role played by each part in enabling effective generalization, with dual normalization being a preferred choice within the module.

\begin{table}[htbp]
\centering
\normalsize
\caption{Ablation studies for Gaussian Kernel on PACS (in $\%$).
The best performance is marked as \textbf{bold} for different backbones, respectively.
GK, TI, and All indicate Gaussian Kernel, Tail Interaction, and All modules, respectively
}
\resizebox{\linewidth}{!}{
\begin{tabular}{l|cccc>{\columncolor{gray!15}}c}
\hline
\toprule[0.5pt]
\specialrule{0em}{1pt}{1pt}
\textbf{Methods}                        & \textbf{Art painting} & \textbf{Cartoon} & \textbf{Photo} & \textbf{Sketch} & \textbf{Avg.}   \\ \specialrule{0em}{1pt}{1pt}\hline \specialrule{0em}{1pt}{1pt}
\multicolumn{1}{l|}{}     & \multicolumn{5}{c}{\textit{ResNet-18}}  \\ \specialrule{0em}{1pt}{1pt} \hline
\specialrule{0em}{1pt}{1pt}
Baseline                      & 77.63\tiny$\pm0.00$                      & 76.77\tiny$\pm0.00$      & 95.85\tiny$\pm0.00$      & 69.50\tiny$\pm0.00$      & 79.94             \\
+ GK                      & 80.38\tiny$\pm0.66$                      & 78.90\tiny$\pm0.82$      & \textbf{96.91}\tiny$\pm0.19$      & 83.13\tiny$\pm1.04$      & 84.83     
           \\
+ TI                      & 79.21\tiny$\pm0.72$                      & 77.03\tiny$\pm0.64$      & 95.91\tiny$\pm0.03$      & 72.09\tiny$\pm0.74$      & 81.06    
           \\
+ All                      & \textbf{83.73}\tiny$\pm0.48$                      & \textbf{81.40}\tiny$\pm0.92$      & 96.89\tiny$\pm0.33$      & \textbf{85.07}\tiny$\pm0.71$      & \textbf{86.77}      
           \\ \specialrule{0em}{1pt}{1pt}\hline \specialrule{0em}{1pt}{1pt}
\multicolumn{1}{l|}{}     & \multicolumn{5}{c}{\textit{GFNet-H-Ti}}  \\ \specialrule{0em}{1pt}{1pt} \hline
\specialrule{0em}{1pt}{1pt}
 Baseline                & \textbf{91.20}\tiny$\pm0.85$        & 82.28\tiny$\pm1.45$     & \textbf{99.03}\tiny$\pm0.14$      & 78.48\tiny$\pm1.28$   &87.75          \\
+ GK                          & 90.63\tiny$\pm0.38$        & 82.78\tiny$\pm1.57$     & 98.78\tiny$\pm0.11$      & 83.17\tiny$\pm1.00$   &88.84 
           \\
+ TI                          & 90.42\tiny$\pm0.62$        & 82.30\tiny$\pm0.87$     & 98.74\tiny$\pm0.06$      & 80.66\tiny$\pm0.87$   &88.03 
           \\
+ All                           & 91.05\tiny$\pm0.14$        & \textbf{83.23}\tiny$\pm1.02$     & 98.67\tiny$\pm0.01$      & \textbf{85.01}\tiny$\pm0.47$   &\textbf{89.49} 
           \\\specialrule{0em}{1pt}{1pt}\hline
\toprule[0.5pt]
\end{tabular}}
\label{tab:ab2}
\vspace{-10pt}
\end{table}

We also conduct ablation studies on our models that utilize the Gaussian Kernel in the frequency restriction module.
The results, shown in Table \ref{tab:ab2}, demonstrate that incorporating either the Gaussian Kernel-based frequency restriction module or the Tail Interaction module improves the generalization performance of the model compared to the baseline. 
However, using both modules simultaneously in the entire network performs better than using only one separately. 
The experimental results indicate that both the frequency restriction module based on Gaussian Kernel and the Tail Interaction module can improve the generalization performance of the model and that the combination achieves more robust learning of domain-invariant representations during training.

\begin{table}[htbp]
\vspace{-10pt}
\centering
\caption{DG accuracy ($\%$) using five fixed and random filter severity levels on Digits-DG.
The backbone is GFNet-H-Ti}
\resizebox{\linewidth}{!}{
\begin{tabular}{l|cccc|>{\columncolor{gray!15}}c}
\specialrule{0em}{1pt}{1pt}\hline\toprule[0.5pt]\specialrule{0em}{1pt}{1pt}
           & \textbf{MNIST}     & \textbf{MNIST-M}     & \textbf{SVHN}     & \textbf{SYN}     & \textbf{Avg.}  \\ \specialrule{0em}{1pt}{1pt}\hline\specialrule{0em}{1pt}{1pt}
% DeepAll    & 97.74\tiny$\pm0.24 & 80.90\tiny$\pm1.20 & 85.15\tiny$\pm1.01 & 96.86\tiny$\pm0.07 & 90.16 \\
Severity=1 & 97.89\tiny$\pm0.06$ & 85.45\tiny$\pm0.49$ & 85.54\tiny$\pm0.61$ & 96.85\tiny$\pm0.17$ & 91.43 \\
Severity=2 & 97.89\tiny$\pm0.07$ & 85.38\tiny$\pm0.53$ & \textbf{85.56}\tiny$\pm0.60$ & 96.84\tiny$\pm0.18$ & 91.42 \\
Severity=3 & 97.89\tiny$\pm0.06$ & 85.44\tiny$\pm0.48$ & \textbf{85.56}\tiny$\pm0.65$ & 96.79\tiny$\pm0.12$ & 91.42 \\
Severity=4 & 97.77\tiny$\pm0.16$ & \textbf{89.37}\tiny$\pm0.33$ & 83.65\tiny$\pm0.95$ & 96.89\tiny$\pm0.18$ & 91.92 \\
Severity=5 & 97.77\tiny$\pm0.19$ & 89.20\tiny$\pm0.33$ & 83.67\tiny$\pm0.72$ & 96.88\tiny$\pm0.19$ & 91.88 \\ \specialrule{0em}{1pt}{1pt}\hline\specialrule{0em}{1pt}{1pt}
Random     & \textbf{97.95}\tiny$\pm0.10$ & 88.87\tiny$\pm0.49$ & 85.31\tiny$\pm0.60$ & \textbf{96.92}\tiny$\pm0.16$ & \textbf{92.26} \\ \specialrule{0em}{1pt}{1pt}\hline\toprule[0.5pt]\specialrule{0em}{1pt}{1pt}
\end{tabular}
}
\label{tab:fixhp}
\end{table}

\vspace{-10pt}
\begin{table}[htbp]
\centering
\caption{DG accuracy ($\%$) with a fixed value of $\alpha$ or $\beta$ and random values of the other in scaling amplitude and phase on Digits-DG.
The backbone is GFNet-H-Ti}
%\rowcolors{1}{blue!20}{blue!10}
\resizebox{\linewidth}{!}{
\begin{tabular}{l|cccc>{\columncolor{gray!15}}c}
\specialrule{0em}{1pt}{1pt}\hline\toprule[0.5pt]\specialrule{0em}{1pt}{1pt}
      & \textbf{MNIST} & \textbf{MNIST-M} & \textbf{SVHN} & \textbf{SYN} & \textbf{Avg.} \\ \specialrule{0em}{1pt}{1pt}\hline\specialrule{0em}{1pt}{1pt}
      \rowcolor{blue!10}
$\alpha=0.6$  & 97.83\tiny$\pm$0.15          & 87.71\tiny$\pm$0.62            & 85.18\tiny$\pm$0.45         & 96.92\tiny$\pm$0.22       & 91.91         \\
$\beta=0.6$  & 97.99\tiny$\pm$0.19               &  87.82\tiny$\pm$0.38                &84.39\tiny$\pm$0.64               & 96.64\tiny$\pm$0.13             & 91.71              \\\rowcolor{blue!10}
$\alpha=0.7$  & 97.92\tiny$\pm$0.11          & 88.22\tiny$\pm$0.61            & 85.04\tiny$\pm$0.52         & 96.93\tiny$\pm$0.18       & 92.03         \\
$\beta=0.7$  &   97.89\tiny$\pm$0.21             &  88.08\tiny$\pm$0.42                &   85.16\tiny$\pm$0.80            & 96.73\tiny$\pm$0.23             & 91.96              \\\rowcolor{blue!10}
$\alpha=0.8$  & 97.86\tiny$\pm$0.15          & 88.59\tiny$\pm$0.59           & 85.00\tiny$\pm$0.73        & 96.85\tiny$\pm$0.11       & 92.07         \\
$\beta=0.8$  & 97.90\tiny$\pm$0.11               & 88.00\tiny$\pm$0.55                 &  \textbf{85.41}\tiny$\pm$0.47             &  96.79\tiny$\pm$0.23            &  92.03             \\\rowcolor{blue!10}
$\alpha=0.9$  & 97.77\tiny$\pm$0.16          & 88.93\tiny$\pm$0.47           & 84.69\tiny$\pm$0.74        & 96.85\tiny$\pm$0.17        & 92.06         \\
$\beta=0.9$  &  97.96\tiny$\pm$0.16              &  87.77\tiny$\pm$0.70                & 85.36\tiny$\pm$0.85              &  \textbf{97.04}\tiny$\pm$0.22            &  92.03             \\\rowcolor{blue!10}
$\alpha=1.0$  & 97.77\tiny$\pm$0.18         & \textbf{88.99}\tiny$\pm$0.45            & 84.24\tiny$\pm$0.22         & 96.87\tiny$\pm$0.20        & 91.96         \\
$\beta=1.0$  & \textbf{98.00}\tiny$\pm$0.20               &  86.55\tiny$\pm$0.86                &  84.76\tiny$\pm$0.89             & 96.82\tiny$\pm$0.23             & 91.53              \\ \specialrule{0em}{1pt}{1pt}\hline\specialrule{0em}{1pt}{1pt}
Random &  97.95\tiny$\pm$0.10              &   88.87\tiny$\pm$0.49               &85.31\tiny$\pm$0.60               & 96.92\tiny$\pm$0.16             &  \textbf{92.26}             \\  \specialrule{0em}{1pt}{1pt}\hline\toprule[0.5pt]\specialrule{0em}{1pt}{1pt}
\end{tabular}}
\label{tab:fixalpha}
\end{table}

\begin{table}[htbp]
\centering
\caption{DG accuracy ($\%$) using different unit sizes in Tail Interaction on Digits-DG.
The backbone is GFNet-H-Ti}
\resizebox{\linewidth}{!}{
\begin{tabular}{l|cccc|>{\columncolor{gray!15}}c}
\specialrule{0em}{1pt}{1pt}\hline\toprule[0.5pt]\specialrule{0em}{1pt}{1pt}
           & \textbf{MNIST}     & \textbf{MNIST-M}     & \textbf{SVHN}     & \textbf{SYN}     & \textbf{Avg.}  \\ \specialrule{0em}{1pt}{1pt}\hline\specialrule{0em}{1pt}{1pt}
% DeepAll    & 97.74\tiny$\pm0.24 & 80.90\tiny$\pm1.20 & 85.15\tiny$\pm1.01 & 96.86\tiny$\pm0.07 & 90.16 \\
Size=16 & 97.87\tiny$\pm$0.11 & 85.40\tiny$\pm$0.52 & \textbf{85.52}\tiny$\pm$0.49 & 96.85\tiny$\pm$0.16 & 91.41 \\
Size=32 & 97.78\tiny$\pm$0.07 & 87.47\tiny$\pm$0.38 & 84.62\tiny$\pm$0.84 & 96.89\tiny$\pm$0.15 & 91.69 \\
Size=64 & \textbf{97.95}\tiny$\pm$0.10 & \textbf{88.87}\tiny$\pm$0.49 & 85.31\tiny$\pm$0.60 & 96.92\tiny$\pm$0.16 & \textbf{92.26} \\
Size=128 & 97.84\tiny$\pm$0.14 & 88.42\tiny$\pm$0.60 & 85.21\tiny$\pm$0.75 & 96.81\tiny$\pm$0.19 & 92.07 \\
Size=256 & 97.79\tiny$\pm$0.15 & 87.63\tiny$\pm$0.45 & 85.19\tiny$\pm$0.62 & 96.91\tiny$\pm$0.20 & 91.88 \\ 
Size=512 & 97.81\tiny$\pm$0.12 & 87.69\tiny$\pm$0.58 & 85.09\tiny$\pm$0.84 & \textbf{96.93}\tiny$\pm$0.23 & 91.88 \\
 \specialrule{0em}{1pt}{1pt}\hline\toprule[0.5pt]\specialrule{0em}{1pt}{1pt}
\end{tabular}}
\label{tab:unitsize}
\end{table}

\begin{table}[htbp]
\centering
%\vspace{-pt} 
%\setlength{\belowcaptionskip}{-0.3cm}
\caption{DG accuracy ($\%$) using different Gaussian Kernel sizes in the frequency restriction module on Digits-DG.
The backbone is GFNet-H-Ti}
\resizebox{\linewidth}{!}{
\begin{tabular}{l|cccc|>{\columncolor{gray!15}}c}
\specialrule{0em}{1pt}{1pt}\hline\toprule[0.5pt]\specialrule{0em}{1pt}{1pt}
           & \textbf{MNIST}     & \textbf{MNIST-M}     & \textbf{SVHN}     & \textbf{SYN}     & \textbf{Avg.}  \\ \specialrule{0em}{1pt}{1pt}\hline\specialrule{0em}{1pt}{1pt}
% DeepAll    & 97.74\tiny$\pm0.24 & 80.90\tiny$\pm1.20 & 85.15\tiny$\pm1.01 & 96.86\tiny$\pm0.07 & 90.16 \\
Size=7 & 97.92\tiny$\pm$0.14 & 86.58\tiny$\pm$0.50 & 84.93\tiny$\pm$0.71 & 96.85\tiny$\pm$0.16 & 91.57 \\
Size=21 & \textbf{97.98}\tiny$\pm$0.09 & 87.90\tiny$\pm$0.45 & 84.72\tiny$\pm$0.56 & 96.56\tiny$\pm$0.20 & 91.79 \\
Size=35 & 97.79\tiny$\pm$0.17 & 87.80\tiny$\pm$0.66 & 84.91\tiny$\pm$0.81 & 96.86\tiny$\pm$0.18 & 91.84 \\
Size=57 & 97.95\tiny$\pm$0.20 & 88.00\tiny$\pm$0.52 & \textbf{85.56}\tiny$\pm$0.74 & \textbf{97.01}\tiny$\pm$0.15 & 92.13 \\
Size=63 & 97.95\tiny$\pm$0.10 & \textbf{88.87}\tiny$\pm$0.49 & 85.31\tiny$\pm$0.60 & 96.92\tiny$\pm$0.16 & \textbf{92.26} \\ 
Size=67 & 97.92\tiny$\pm$0.18 & 88.73\tiny$\pm$0.61 & 85.43\tiny$\pm$0.48 & 96.88\tiny$\pm$0.20 & 92.24 \\
 \specialrule{0em}{1pt}{1pt}\hline\toprule[0.5pt]\specialrule{0em}{1pt}{1pt}
\end{tabular}
}
\label{tab:kersize}
\end{table}

\subsection{Further Analysis}
\label{sec:fur}
In this section, we present detailed experimental results about the effect of some crucial hyper-parameters of Figure 4 in the main paper.
The experiments are conducted on Digits-DG with GFNet-H-Ti as the backbone.

(1) Impact of high-pass filter severity.
The results presented in Table \ref{tab:fixhp} show that the model performs better when using random severity levels for the high-pass filter operation compared to when using one of the five severity levels specifically. 
Based on this, we believe that an adaptive scheme that adjusts the severity level of the filter for samples from different domains can be more effective and reasonable than using fixed or random severity levels.

(2) Impact of $\alpha$ and $\beta$.
Since using a filter transfers the image from the spatial domain to the frequency domain, we take this opportunity to use the amplitude and phase information contained in the spectrum for noise reduction.
Table \ref{tab:fixalpha} shows the results of varying the strength $\alpha$ or $\beta$ from 0.6 to 1.0, and the value of the other one is random. 
When the target domain is less dissimilar to the source domains, such as MNIST and SYN, the results do not vary too much with different $\alpha$ and $\beta$. 
Our model still shows a moderate sensitivity to $\alpha$ and $\beta$ when the target domain exhibits a more extensive domain gap than the sources, \eg, MNIST-M and SVHN.

(3) Impact of the unit size of Tail Interaction.
The Tail Interaction unit offers the flexibility of manual adjustment to achieve a small computational complexity, which is linear in the number of pixels. 
Our experiments, as shown in Table \ref{tab:unitsize}, indicate that even with a small interaction unit size, the Tail Interaction effectively improves the generalization performance of the model.

(4) Impact of Gaussian Kernel size.
According to the results presented in Table \ref{tab:kersize}, a small Gaussian Kernel employed as the low frequency filter can effectively preserve high frequency components as domain-invariant representations and efficiently generate augmented images.

\end{document}